\documentclass[11pt]{article}

\usepackage[final]{acl}
\usepackage{tabularx}
\usepackage{booktabs}
\usepackage{array}
\usepackage{booktabs}
\usepackage{pifont}
\usepackage{ragged2e} 
\usepackage{xcolor}   
\usepackage{times}
\usepackage{latexsym}
\usepackage{bm}
\usepackage{amsmath}
\usepackage{amssymb}
\usepackage{subfigure}
\usepackage{booktabs}
\usepackage[normalem]{ulem}
\usepackage[T1]{fontenc}
\usepackage[table]{xcolor} 
\usepackage{multirow}      
\usepackage[utf8]{inputenc}

\usepackage{microtype}
\usepackage{tabularx}
\usepackage{inconsolata}
\usepackage{silence}
\usepackage{graphicx}
\newcommand{\dec}[1]{{\fontsize{8pt}{7pt}\selectfont\color{black!70}(-#1\%)}}
\newcommand{\pad}{\phantom{\dec{0.0}}}

\title{GS-Quant: Granular Semantic and Generative Structural Quantization for Knowledge Graph Completion}

\author{
\textbf{Qizhuo Xie},
\textbf{Yunhui Liu},
\textbf{Yu Xing},
\textbf{Qianzi Hou},
\textbf{Xudong Jin},
\textbf{Tao Zheng},
\textbf{Tieke He}\thanks{Corresponding author.}
\\State Key Laboratory for Novel Software Technology, Nanjing University, China\\
\{mikumifa, hetieke\}@gmail.com
}
\begin{document}
\maketitle

\begin{abstract} 
Large Language Models (LLMs) have shown immense potential in Knowledge Graph Completion (KGC), yet bridging the modality gap between continuous graph embeddings and discrete LLM tokens remains a critical challenge. 
While recent quantization-based approaches attempt to align these modalities, they typically treat quantization as flat numerical compression, resulting in semantically entangled codes that fail to mirror the hierarchical nature of human reasoning. 
In this paper, we propose GS-Quant, a novel framework that generates semantically coherent and structurally stratified discrete codes for KG entities. 
Unlike prior methods, GS-Quant is grounded in the insight that entity representations should follow a linguistic coarse-to-fine logic. 
We introduce a Granular Semantic Enhancement module that injects hierarchical knowledge into the codebook, ensuring that earlier codes capture global semantic categories while later codes refine specific attributes. 
Furthermore, a Generative Structural Reconstruction module imposes causal dependencies on the code sequence, transforming independent discrete units into structured semantic descriptors. 
By expanding the LLM vocabulary with these learned codes, we enable the model to reason over graph structures isomorphically to natural language generation. 
Experimental results demonstrate that GS-Quant significantly outperforms existing text-based and embedding-based baselines. 
Our code is publicly available at \url{https://github.com/mikumifa/GS-Quant}.
\end{abstract}

\section{Introduction}

Knowledge Graphs (KGs) serve as a cornerstone for symbolic knowledge representation, supporting explicit reasoning and driving progress in applications ranging from recommender systems~\cite{KG4RecSurvey, K-RagRec} and question answering~\cite{GraphRAG, HippoRAG, LLM4KGQASurvey} to the mitigation of hallucination in Large Language Models (LLMs)~\cite{KG4HalluSurvey, ToG}. 
Despite their widespread adoption, KGs suffer from pervasive incompleteness: they encode observed facts while inevitably omitting numerous valid triples~\cite{KGCSurvey, SAT}. 
This limitation necessitates Knowledge Graph Completion (KGC), a task aiming to infer missing links.

Recently, LLMs have fundamentally reshaped the landscape of KGC~\cite{KGLLM} by introducing powerful generative reasoning capabilities. 
Prior approaches broadly fall into two paradigms: text-based and embedding-based. 
Text-based methods~\cite{KICGPT, MPIKGC, MKGL, KGR3, CoLE} linearize KG triples into textual prompts. While this provides explicit reasoning paths, the linearization process shatters the intrinsic graph topology~\cite{SAT} and incurs prohibitive computational costs due to excessive token length~\cite{SSQR}. 
Conversely, embedding-based methods~\cite{KoPA, DIFT, FtG, SLiNT} align rigid, continuous KG embeddings~\cite{TransE, RotatE} with the LLM’s latent space. 
While computationally efficient, a fundamental \textit{modality gap} persists: continuous embeddings are holistic and dense, whereas LLMs operate on discrete, sequential tokens. This mismatch hinders the LLM's ability to effectively interpret and reason over graph structures.

To bridge this gap, recent works like SSQR~\cite{SSQR} and ReaLM~\cite{ReaLM} have explored representing entities as sequences of discrete codes via Vector Quantization~\cite{VQVAE, GTBPE}. 
By converting embeddings into code tokens, these methods align the data format with linguistic sequences. 
However, current quantization approaches in KGC remain suboptimal. 
For example, SSQR derives codes of an entity based on multiple feed-forward projections of its embeddings, resulting in flat and semantically entangled representations.  
The resulting code sequences lack intrinsic semantic logic; they function merely as compressed numerical approximations of a vector, rather than as structured, hierarchical semantic descriptors.
This ignores a critical property of human language and LLM generation: reasoning is inherently hierarchical, proceeding from coarse-grained concepts to fine-grained details~\cite{tenenbaum2011grow, HiTs}. 
Treating quantization as a flat compression problem fails to leverage the autoregressive nature of LLMs, which thrive on structured, causal dependencies.

\begin{figure}
    \centering
    \includegraphics[width=0.49\textwidth]{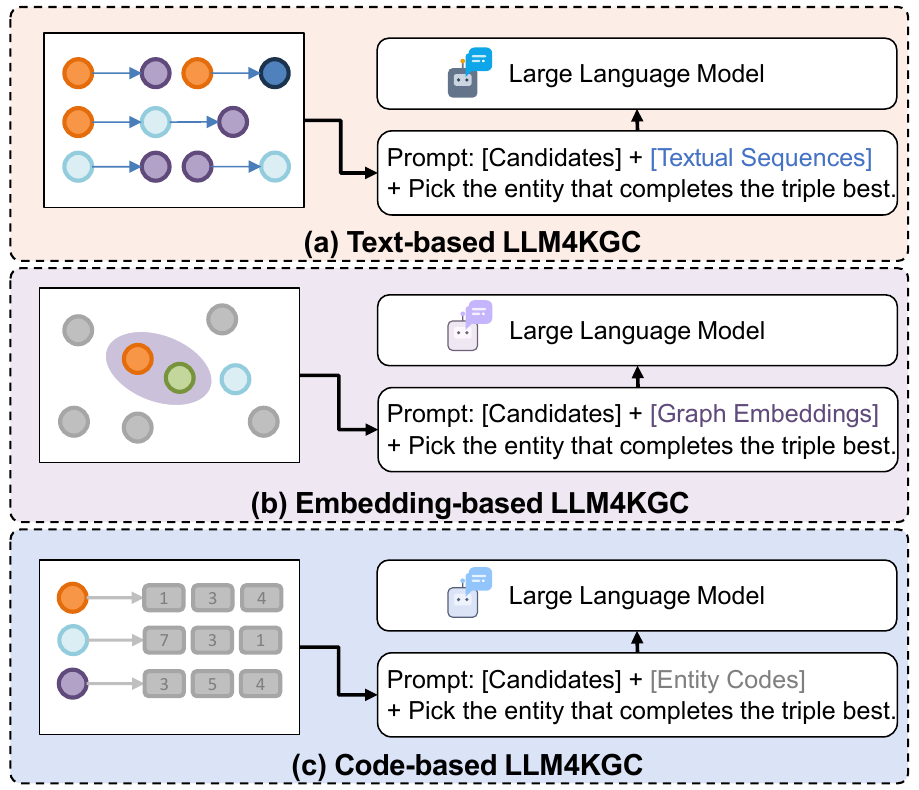}
    \caption{Illustration of LLM4KGC methods. (a) Text-based methods flatten the KG into textual sequences. (b) Embedding-based methods integrating graph embeddings into LLMs. (c) Our code-based method learns quantized codes for better LLM reasoning.}
    \label{fig:llm4kgc_illustration}
\end{figure}

In this paper, we propose GS-Quant, a framework that generates semantically coherent and structurally stratified quantized codes for KG entities. 
GS-Quant is grounded in the insight that an entity's discrete representation should mirror the hierarchical structure of language, i.e., moving from general semantic categories to specific instances. 
Built upon the Residual Quantized Variational Autoencoder (RQ-VAE)~\cite{RQVAE}, our framework introduces two key innovations. 
First, the \textbf{Granular Semantic Enhancement (GSE)} module injects hierarchical knowledge (derived via clustering) into the codebook learning process. 
By forcing different quantization levels to align with a hierarchy tree, GSE ensures that earlier codes capture coarse-grained semantics (e.g., "Person") while later codes refine specific attributes (e.g., "Artist"), effectively creating a semantic coordinate system for the LLM. 
Second, the \textbf{Generative Structural Reconstruction (GSR)} module employs a lightweight GPT-style Transformer decoder to reconstruct the entity and its ancestors from the code sequence. 
This imposes a causal structure on the codes, ensuring that the discrete sequence is not just a set of independent units but a coherent "sentence" that encodes contextual and cross-level interactions.
By treating KG entities as hierarchically structured code sequences, GS-Quant allows LLMs to reason over graph data in a manner isomorphic to natural language generation. 
We expand the LLM's vocabulary with the learned discrete codes and fine-tune the LLM to perform knowledge graph completion.
Experimental results demonstrate that GS-Quant significantly outperforms existing baselines, setting a new standard for LLM-enhanced KGC.

\section{Related Work}
\paragraph{Traditional Embedding-based KGC methods.}Embedding-based KGC methods learn low-dimensional representations of entities and relations under various geometric constraints. Classical translational and multiplicative models include TransE~\cite{TransE}, DistMult~\cite{DistMult}, ComplEx~\cite{Complex}, and RotatE~\cite{RotatE}, while ATTH~\cite{ATTH} and GIE~\cite{GIE} extend them to hyperbolic spaces to better capture hierarchical patterns in KGs.
Beyond shallow embeddings, Graph Neural Networks~(GNNs)~\cite{GNN,DBLP:conf/nips/SongZK23a,DBLP:conf/nips/SongZK23} have been applied to encode relational structures, with applications in various domains~\cite{ConcreteGraph,NewFrontiersinGraphLearning,DBLP:conf/aaai/SongMK24}. To handle relational heterogeneity, RGCN~\cite{RGCN}, WGCN~\cite{WGCN}, and CompGCN~\cite{CompGCN} introduce relation-aware message passing, and KBGAT~\cite{KBGAT} incorporates attention mechanisms to weigh triple importance.
Path-based models such as NBF-Net~\cite{NBF-Net}, RED-GNN~\cite{RED-GNN}, and A*Net~\cite{ANet} exploit multi-hop relational paths to enhance inference and inductive generalization. Despite their efficiency and scalability, these approaches still lack the capacity to capture the rich semantic information of entities.

\paragraph{LLM-based KGC methods. }
Early works like KG-BERT~\cite{kgbert}, KEPLER~\cite{KEPLER}, and CoLE~\cite{CoLE} leverage PLMs to encode triples. Building upon these foundations, current LLM-based approaches primarily diverge into text-based and embedding-based paradigms. 
Text-based methods~\cite{KICGPT, MPIKGC, MKGL, KGR3} serialize triples into natural language prompts, occasionally augmenting them with generated descriptions~\cite{CD, KGFIT}, though this often incurs high token costs and structural information loss~\cite{SAT, SSQR}. 
Conversely, embedding-based methods~\cite{KoPA, DIFT, FtG, SLiNT} inject structural priors by aligning pretrained KG embeddings with the LLM's latent space. 
To further bridge the modality gap between continuous embeddings and discrete text, recent studies explore discrete codes, such as SSQR~\cite{SSQR} using vector quantization and ReaLM~\cite{ReaLM} applying residual quantization. 
Despite their progress, these methods remain limited: code-based approaches largely expand a single entangled embedding, offering weak semantic or structural disentanglement, and SFT-based~\cite{SFT} methods do not bridge the gap between KG embedding space and the LLM’s semantic space. These challenges motivate the need for more structured and semantically coherent discrete representations, which  $\text{GS-Quant}$ framework  aims to provide.

\section{Preliminaries}
\paragraph{Knowledge Graph. }A Knowledge Graph (denoted as $\mathcal{G} = \{\mathcal{E}, \mathcal{R}, \mathcal{T}\}$) can be formally represented as a set of triples in the form of $(h, r, t)\in \mathcal{T}$, where $h, t \in \mathcal{E}$, $r \in \mathcal{R}$. 
The notations $h$ and $t$ denote the head and the tail entity of a triple. 

\paragraph{LLM-based KGC. }In the LLM-based formulation of KGC, we illustrate the task using the tail prediction query $(h,r,?)$ as an example. For a given query, we construct an input to the LLM that combines (i) the query triple, (ii) structural relational context, and (iii) textual information from the KG. The LLM is then required to select the correct tail entity $t$ from a predefined candidate set.

\begin{figure*}[ht!]
    \centering
    \includegraphics[width=1.0\textwidth]{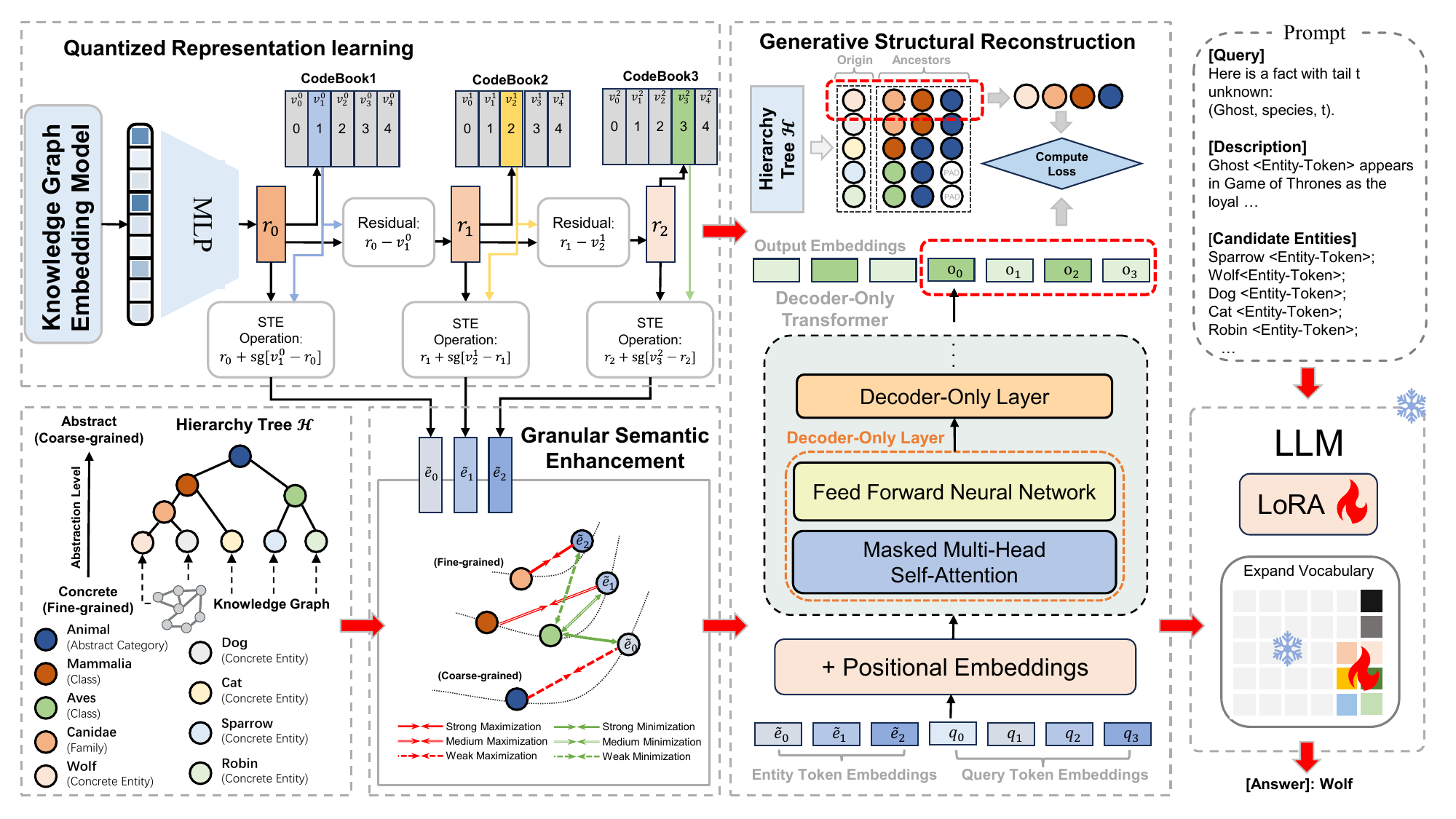}
    \caption{The overview of the $\text{GS-Quant}$ framework. The pipeline first constructs a hierarchy tree. It then discretizes entities using Residual Quantization, optimized by \textbf{Granular Semantic Enhancement} for structural consistency. Subsequently, the \textbf{Generative Structural Reconstruction} module employs a GPT-style Transformer decoder to capture contextual and cross-level interactions. Finally, the learned codes are integrated into LLMs.}
    \label{fig:overview}
\end{figure*}

\section{Methodology}
In this section, we present $\text{GS-Quant}$ (Figure~\ref{fig:overview}). We first encode entities into dense embeddings and construct a semantic tree leveraging hierarchical clustering. 
The embeddings and tree are then utilized to train the residual quantization module to obtain the discrete tokens of the graph. 
This quantization process is jointly optimized using Generative Structural Reconstruction and Granular Semantic Enhancement. 
Subsequently, the learned discrete codes are utilized to expand the LLM's vocabulary, and the model with the augmented vocabulary is then fine-tuned for the KGC task.

\subsection{Quantized Codes Learning for KG}
\subsubsection{Basic Codes Learning}
\paragraph{KG Encoding.}
For the structural component, a knowledge graph embedding model (e.g., RotatE~\cite{RotatE}) is employed as the backbone to generate the relational embedding $\mathbf{s}_x^{\mathcal{G}} \in \mathbb{R}^d$.

For the textual component, we utilize both the entity name and description. Specifically, we encode them independently using a pretrained PLM and concatenate their respective truncated feature representations to form the textual embedding $\mathbf{s}_x^T \in \mathbb{R}^d$, thereby ensuring dimensional alignment with $\mathbf{s}_x^{\mathcal{G}}$. The final representation $\mathbf{s}_x$ is obtained via a weighted fusion:
\begin{equation}
    \mathbf{s}_x = \rho\, \mathbf{s}_x^{\mathcal{G}}
    + (1 - \rho)\, \mathbf{s}_x^T,
\end{equation}
where $\rho \in [0, 1]$ controls the relative balance between relational and textual contributions.

\paragraph{Residual Quantization (RQ).}
We adopt the residual quantization mechanism~\cite{RQVAE} to encode the enriched entity embedding $\mathbf{s}$, which can be either $\mathbf{s}_h$ or $\mathbf{s}_t$, to learn a latent representation
$\mathbf{z} = \text{MLP}(\mathbf{s})$, which serves as the
initial residual at the 0-th level, denoted as $\mathbf{r}_0 = \mathbf{z}$. At each
quantization level $l$, there is a codebook
$\mathbf{C}^{l} = \{ \mathbf{v}^{l}_k \}_{k=1}^{K}$, where $\mathbf{v}^{l}_k$ is the $k$-th vector
in the codebook at level $l$ and $K$ is the codebook size.

The initial residual $\mathbf{r}_0$ is then used to find the index of the
nearest embedding in $\mathbf{C}^{0}$, given by
$\mathbf{v}_0 = \arg\min_{k}\| \mathbf{r}_0 - \mathbf{v}^{0}_k \|_2$. At each level
$l$, the residual update follows the rule
$c_l = \arg\min_{k}\| \mathbf{r}_l - \mathbf{v}^{l}_k \|_2$, and the residual is iteratively
updated as $\mathbf{r}_{l+1} = \mathbf{r}_l - \mathbf{v}^{l}_{c_l}$. This recursive residual
approximation ultimately generates a quantization code tuple
$\mathcal{I} = \{ c_i \}_{i=0}^{m-1}$, where $m$
is the maximum level of RQ.

\paragraph{Residual Quantization Regularization.}
Following generating the code tuple $\mathcal{I}$, we apply a
stop-gradient based commitment loss to stabilize codebook learning.
The residual quantization loss is:

{\small
\begin{equation}
\label{eq:rq}
\begin{split}
    \mathcal{L}_{\text{Q}} &= \sum_{l=0}^{m-1} \Big(
    \| \operatorname{sg}[\mathbf{r}_l] - \mathbf{v}^{l}_{c_l} \|_2^2
    + \alpha \| \mathbf{r}_l - \operatorname{sg}[\mathbf{v}^{l}_{c_l}] \|_2^2
    \Big)
\end{split}
\end{equation}
}

where $\operatorname{sg}[\cdot]$ denotes the stop-gradient operator, which prevents gradient updates for the quantized embeddings during backpropagation.
The first term in $\mathcal{L}_{\text{Q}}$ ensures that the trainable codebook vectors $\mathbf{v}^{l}_{c_l}$ are close to the corresponding residuals $\mathbf{r}_l$. Meanwhile, the second term, weighted by the hyperparameter $\alpha$, constrains the residuals to remain close to the selected codebook entries. 

Eq.~\eqref{eq:rq} alone is insufficient to ensure semantically meaningful codebooks. We introduce additional components in the following section.

\subsubsection{Granular Semantic Enhancement}
In KGC, accurate prediction often requires hierarchical inference. For instance, in the query (Christmas, month, ?), the model should ideally identify the coarse-grained semantic category of "month" before narrowing down to the specific entity "December". However, traditional residual vector quantization is insufficient for this purpose. While it naturally exhibits a numerical hierarchy due to the recursive residual subtraction $\mathbf{r}_{l+1} = \mathbf{r}_l - \mathbf{v}^{l}_{c_l}$, it lacks explicit alignment with distinct semantic granularity levels. Each quantization layer merely encodes the numerical residual of the previous one based on Euclidean distance. The resulting codebooks often entangle semantics across layers, failing to explicitly follow the coarse-to-fine semantic hierarchy required for effective knowledge graph completion.\par

To address this, we first perform hierarchical clustering (e.g., agglomerative clustering~\cite{agglomerativeclustering}) on entity semantic representations to generate a hierarchy tree $\mathcal{H}$. 
Let $E$ denote a batch of entities. For each $\mathbf{e} \in E$, let $\boldsymbol{\mu}_{\mathbf{e}}$ be its corresponding cluster centroid. We define a differentiable surrogate for each quantized output of $\mathbf{e}$ as $\mathcal{V}_e=\{\tilde{\mathbf{v}}_i\}_{i=0}^{m-1}$, formulated as $\tilde{\mathbf{v}}_i = \mathbf{r}_i + \operatorname{sg}[\mathbf{v}^i_{c_i} - \mathbf{r}_i]$, where $\mathbf{r}i$ denotes the residual representation at step $i$, and $\mathbf{v}^i_{c_i}$ is the selected codebook vector from the $i$-th codebook. The stop-gradient operator $\operatorname{sg}[\cdot]$ ensures stable optimization by allowing gradients to flow through the residual pathway while preventing direct updates to the discrete codebook selection.

\paragraph{Coarse-to-Fine Alignment.} GSE calibrates the quantized output $\mathcal{V}_e$ to capture coarse-to-fine semantic granularity. The loss is defined  as:
\begin{equation}
\begin{split}
    \mathcal{L}_1 &= -\frac{1}{|E|} \sum_{\mathbf{e} \in E} \sum_{i=0}^{m-1} \frac{\lambda_1^{i+1}}{m} \\
    &\quad \times \log \frac{\exp(\tilde{\mathbf{v}}_i \cdot \boldsymbol{\mu}_{\mathbf{e}} / \tau)}{\sum_{\mathbf{e}' \in E} \exp(\tilde{\mathbf{v}}_i \cdot \boldsymbol{\mu}_{\mathbf{e}'} / \tau)}.
\end{split}
\end{equation}
where $\tau$ is the temperature hyperparameter and $\lambda_1 \in (0,1)$ controls an exponentially decaying weight across layers.
This design biases earlier layers toward coarse, global semantics, while allowing deeper layers to focus on finer-grained distinctions.

\paragraph{Hierarchical Separability. } GSE distinguishes each vector in $\mathcal{V}_e$ from its neighbors in the hierarchy tree $\mathcal{H}$. Let $\mathcal{N}_e$ denote the set of neighbor centroids for entity $e$. This loss is formulated as:
\begin{equation}
\begin{split}
    \mathcal{L}_2 &= \frac{1}{|E|} \sum_{\mathbf{e} \in E} \sum_{i=0}^{m-1} \frac{\lambda_2^{m-i}}{m \cdot |\mathcal{N}_{\mathbf{e}}|} \\
    &\quad \times \sum_{\mathbf{n} \in \mathcal{N}_{\mathbf{e}}} \log \frac{\exp(\tilde{\mathbf{v}}_i \cdot \mathbf{n} / \tau)}{\sum_{\mathbf{e}' \in E} \exp(\tilde{\mathbf{v}}_i \cdot \boldsymbol{\mu}_{\mathbf{e}'} / \tau)}.
\end{split}
\end{equation}
where $\lambda_2 \in (0,1)$ applies a reverse decay that progressively strengthens constraints in deeper layers. This encourages fine-grained separability while allowing higher layers to remain semantically compact. The total GSE objective is $
    \mathcal{L}_{\text{GSE}} = \mathcal{L}_1 + \mathcal{L}_2$.
\subsubsection{Generative Structural Reconstruction}

For the code tuple $\mathcal{I}$, its rich combinatorial semantic capacity is essential for downstream LLM generation tasks. Different code combinations ought to encode distinct semantic variations, enabling the model to support complex generative reasoning. 

To achieve this, we introduce Generative Structural Reconstruction. We employ a standard Transformer decoder to treat the tuple $\mathcal{I}$ as a structured semantic sequence. Formally, we construct a sequence of learnable query embeddings $\mathcal{Q} = \{ \mathbf{q}_i \}_{i=0}^{L}$ and concatenate them with the differentiable surrogate $\mathcal{V}_e$ as the input. Through causal self-attention, the decoder produces reconstruction outputs $\{ \mathbf{o}_i \}_{i=0}^{L}$ aligned with multi-granular targets: the leading query $\mathbf{q}_0$ is tasked with recovering the intrinsic entity embedding $\mathbf{s}$, while subsequent queries $\{ \mathbf{q}_i \}_{i=1}^{L}$ reconstruct the corresponding ancestors $\{\mathbf{h}_i\}$ in the semantic hierarchy $\mathcal{H}$ to enforce global consistency. The overall GSR objective is formulated as:
\begin{equation}
\begin{split}
    \mathcal{L}_{\text{GSR}} &= \|\tilde{\mathbf{o}}_{0} - \mathbf{s}\|_2^2 + \lambda_{s} \|\tilde{\mathbf{o}}_{1} - \mathbf{h}_{0}\|_2^2 \\
    &\quad + \lambda_{h} \sum_{i=2}^{L} \|\tilde{\mathbf{o}}_{i} - \mathbf{h}_{i-1}\|_2^2 .
\end{split}
\end{equation}
where $\lambda_{s}$ weights the reconstruction of the semantic target $\mathbf{h}_{0}$, 
and $\lambda_{h}$ controls the strength of reconstructing the remaining hierarchical semantic targets 
$\{\mathbf{h}_i\}_{i=1}^{L}$. 
In practice, we assign a significantly smaller value to $\lambda_{s}$ compared to $\lambda_{h}$, considering that $\mathbf{h}_{0}$ has been integrated into the GSE module.

Importantly, This design not only significantly enhances the combinatorial semantic capacity of the discrete codes, but also firmly anchors them to both fine-grained and coarse-grained semantic targets, thereby ensuring coarse-to-fine semantic alignment within the quantized representation.

\subsubsection{Full Quantization Learning}
The final training objective is the summation of the three loss components:
\begin{equation}
    \mathcal{L}_{\text{total}} = \mathcal{L}_{\text{Q}} + \mathcal{L}_{\text{GSE}} + \mathcal{L}_{\text{GSR}}.
\end{equation}
To ensure efficient utilization of the discrete latent space across all quantization levels, We select the model checkpoint based on the highest Codebook Entropy. Let $p_k^m$ denote the empirical activation frequency of the $k$-th code within the $m$-th codebook. The Codebook Entropy $\mathcal{Y}$ is calculated as:
\begin{equation}
    \mathcal{Y} = - \frac{1}{M} \sum_{m=1}^{M} \sum_{k=1}^{K} p_k^m \log p_k^m.
\end{equation}
Maximizing $\mathcal{Y}$ encourages the model to fully exploit the inherent expressive capacity of the dictionary, as the entropy attains its maximum value when all codes are utilized with equal probability.

\subsection{Fine-tuning LLMs with Quantized Codes}
To seamlessly integrate the learned quantized representations, we explicitly extend the LLM's vocabulary with tokens corresponding to the codebook entries. To facilitate efficient adaptation, we freeze the original model parameters and update only specific components: (1) the embeddings of the newly added code tokens, allowing them to align with the LLM's generation dynamics; and (2) the adapter matrices introduced via Low-Rank Adaptation (LoRA)~\cite{LoRA} in the attention and feed-forward layers. This dual approach ensures the effective injection of rich domain knowledge while preserving the LLM's general capabilities.

\section{Experiments}

\subsection{Baselines and Evaluation Metrics}
We compare the $\text{GS-Quant}$ framework against three distinct categories of baseline methods: (1) Embedding-based methods, including TransE~\cite{TransE}, ComplEx~\cite{Complex}, RotatE~\cite{RotatE}, TuckER~\cite{TuckER}, CompGCN~\cite{CompGCN}; (2) Text-based methods, including KG-BERT~\cite{kgbert}, MEM-KGC~\cite{MEM-KGC}, CoLE~\cite{CoLE}; and (3) LLM-based methods including, KICGPT~\cite{KICGPT}, DIFT~\cite{DIFT}, KG-FIT~\cite{KGFIT}, MKGL~\cite{MKGL}, SSQR~\cite{SSQR}.

We utilize the widely adopted evaluation metrics, namely Hits@$k$~($k=1,3,10$) and MRR. Hits@$k$ measures the proportion of query triples which the ground truth entities are ranked within the top-$k$ position. MRR measures the mean reciprocal rank for each ground truth entities. In our framework, the finetuned LLM selects an entity as the answer from the ranking list of candidates. To assess its performance and make the results comparable to existing work, we move the selected entity to the top of the ranking list, and other candidates remain unchanged. Therefore, the gains in Hit@1 and MRR best highlight the practical value of our approach, as they reflect the model's ability to provide the correct answer immediately. Implementation details and the measures taken to ensure a fair comparison are described in Appendix~\ref{sec:implementation}.

\subsection{Main Results}

\begin{table*}[t]
\setlength{\tabcolsep}{1.5pt}
  \label{tab:lp_merged}
  \begin{tabular}{lcccccccc}
    \toprule
    \multirow{2}{*}{\parbox[c]{3cm}{\centering \textbf{Model}}} 
    & \multicolumn{4}{c}{\textbf{WN18RR}} 
    & \multicolumn{4}{c}{\textbf{FB15k237}} \\
    \cmidrule(lr){2-5} \cmidrule(lr){6-9}
    & \textbf{MRR} & \textbf{Hits@1} & \textbf{Hits@3} & \textbf{Hits@10} & \textbf{MRR} & \textbf{Hits@1} & \textbf{Hits@3} & \textbf{Hits@10} \\
    \midrule
    \rowcolor[gray]{0.9} \multicolumn{9}{c}{\textbf{Embedding-based Methods}} \\
    TransE~\cite{TransE}  & 0.243 & 0.043 & 0.441 & 0.532 & 0.279 & 0.198 & 0.376 & 0.441 \\
    ComplEx~\cite{Complex} & 0.440 & 0.410 & 0.460 & 0.510 & 0.247 & 0.158 & 0.275 & 0.428 \\
    RotatE~\cite{RotatE}  & 0.476 & 0.428 & 0.492 & 0.571 & 0.338 & 0.241 & 0.375 & 0.533 \\
    TuckER~\cite{TuckER}  & 0.470 & 0.443 & 0.482 & 0.526 & 0.358 & 0.266 & 0.394 & 0.544 \\
    CompGCN~\cite{CompGCN}  & 0.479 & 0.443 & 0.494 & 0.546 & 0.355 & 0.264 & 0.390 & 0.535 \\
    \midrule
    \rowcolor[gray]{0.9} \multicolumn{9}{c}{\textbf{Text-based Methods}} \\
    KG-BERT~\cite{kgbert}  & 0.216 & 0.041 & 0.302 & 0.524 & -     & -     & -     & 0.420 \\
    MEM-KGC~\cite{MEM-KGC}  & 0.557 & 0.475 & 0.604 & 0.704 & 0.346 & 0.253 & 0.381 & 0.531 \\
    CoLE~\cite{CoLE}     & 0.593 & 0.538 & 0.616 & 0.701 & 0.389 & 0.294 & 0.429 & 0.572 \\
    \midrule
    \rowcolor[gray]{0.9} \multicolumn{9}{c}{\textbf{LLM-based Methods}} \\
    KICGPT~\cite{KICGPT}         & 0.564 & 0.478 & 0.612 & 0.677 & 0.412 & 0.327 & 0.448 & 0.581 \\
    DIFT~\cite{DIFT}         & \underline{0.617} & \underline{0.569} & \underline{0.638} & \underline{0.708} & \underline{0.439} & \underline{0.364} & \underline{0.468} & 0.586 \\
    KG-FIT~\cite{KGFIT}  & 0.553 & 0.488 & - & 0.695 & 0.362 & 0.275 & - & 0.572 \\
    MKGL~\cite{MKGL}  & 0.552 & 0.500 & 0.577 & 0.656 & 0.415 & 0.325 & 0.454 & \underline{0.591} \\
    SSQR~\cite{SSQR}    & 0.603 & 0.553  & 0.627 & 0.692 & 0.428 & 0.349 & 0.459  & 0.583 \\
    \midrule
    \textbf{$\text{GS-Quant}$}  & \textbf{0.635} & \textbf{0.594} & \textbf{0.649} & \textbf{0.712} & \textbf{0.455} & \textbf{0.386} & \textbf{0.479} & \textbf{0.592} \\
    \bottomrule
  \end{tabular}
    \caption{Comparison of KGC results on WN18RR and FB15k-237 datasets. 
    The results are categorized into Embedding-based, Text-based, and LLM-based methods for fair comparison. 
    The best results \textbf{boldfaced} and the second-best are \underline{underlined}.}

  \label{tab:results}
\end{table*}

\begin{table}[t]
\centering
\setlength{\tabcolsep}{1.5pt} 
\resizebox{\linewidth}{!}{ 
    \begin{tabular}{lcccc}
    \toprule
    \textbf{Settings} & \textbf{MRR} & \textbf{Hits@1} & \textbf{Hits@3} & \textbf{Hits@10} \\

    \midrule
    \multicolumn{5}{c}{\textbf{FB15k-237}} \\
    \midrule
    \textbf{Ours} & \textbf{0.455} \pad & \textbf{0.386} \pad & \textbf{0.479} \pad & \textbf{0.592} \pad\\
    w/o $\mathcal{L}_1$ & 0.450 \dec{0.5} & 0.377 \dec{0.9} & 0.477 \dec{0.2} & 0.590 \dec{0.2} \\
    w/o $\mathcal{L}_2$ & 0.450 \dec{0.5} & 0.379 \dec{0.7} & 0.476 \dec{0.3} & 0.589 \dec{0.3} \\
    w/o $\mathcal{L}_{GSR}$ & 0.448 \dec{0.7} & 0.375 \dec{1.1} & 0.476 \dec{0.3} & 0.590 \dec{0.2} \\
    w/o Code & 0.404 \dec{5.1} & 0.303 \dec{8.3} & 0.457 \dec{2.2} & 0.584 \dec{0.8} \\

    \midrule
    \multicolumn{5}{c}{\textbf{WN18RR}} \\ 
    \midrule
    \textbf{Ours} & \textbf{0.635} \pad & \textbf{0.594} \pad & \textbf{0.649} \pad & \textbf{0.712} \pad \\
    w/o $\mathcal{L}_1$ & 0.629 \dec{0.5} & 0.587 \dec{0.6} & 0.646 \dec{0.4} & 0.711 \dec{0.1} \\
    w/o $\mathcal{L}_2$ & 0.625 \dec{0.9} & 0.577 \dec{1.6} & 0.646 \dec{0.4} & 0.710 \dec{0.2} \\
    w/o $\mathcal{L}_{GSR}$ & 0.627 \dec{0.7} & 0.585 \dec{0.8} & 0.642 \dec{0.8} & 0.710 \dec{0.2} \\
    w/o Code & 0.607 \dec{2.7} & 0.541 \dec{5.2} & 0.640 \dec{1.0} & 0.708 \dec{0.4} \\
    \bottomrule
    \end{tabular}
}
\caption{Ablation study on FB15k-237 and WN18RR.}
\label{tab:ablation_vertical_percent}
\end{table}

Table~\ref{tab:results} summarizes the performance of the $\text{GS-Quant}$ framework. Specifically, compared to the strongest LLM-based baseline, GS-Quant achieves significant improvements across both datasets. On WN18RR, we observe a gain of approximately \textbf{1.7\%} in MRR and \textbf{2.4\%} in Hits@1, indicating more accurate top-ranked predictions. Similarly, on the FB15k-237 dataset, GS-Quant surpasses the strongest LLM-based baseline with improvements of \textbf{1.6\%} in MRR and \textbf{2.2\%} in Hits@1, demonstrating its robustness across datasets.

Notably, although we utilize the exact same candidate sets and instruction templates as DIFT in our experiments, GS-Quant consistently outperforms DIFT, particularly in Hits@1 on both datasets. This phenomenon strongly suggests that the quantized representations employed in GS-Quant are significantly more effective for LLM reasoning than the continuous embeddings used in DIFT. 

Furthermore, GS-Quant shows clear advantages over SSQR, which also employs quantized representations. This performance boost is attributed to our codebook design, which not only ensures explicit alignment with distinct semantic granularity levels but also encourages the discrete units to capture contextual and cross-level interactions. These structured features make the representations inherently more aligned with the LLM's understanding and more effective for downstream tasks.

\subsection{Ablation Studies}
To assess the effectiveness of each component in our framework, we conducted a series of ablation studies, with results presented in Table~\ref{tab:ablation_vertical_percent}. All ablation settings share the same training configuration, data splits, and optimization hyperparameters. Here, w/o $\mathcal{L}_1$ and w/o $\mathcal{L}_2$ each remove one of the two components that constitute the GSE module. w/o $\mathcal{L}_{\text{GSR}}$ disables the GSR while still retaining the use of quantized code tokens. Finally, w/o Code eliminates all quantized tokens from the vocabulary, preventing the LLM from leveraging any discrete semantic units produced by the quantization process. Several key observations can be seen:
(1) Both $\mathcal{L}_1$ and $\mathcal{L}_2$ are essential for GSE. Removing either $\mathcal{L}_1$ or $\mathcal{L}_2$ leads to consistent performance drops on both datasets, indicating that the two objectives jointly support the formation of coherent multi-granular semantics
(2) GSR further boosts the utility of multi-level quantization. Disabling GSR weakens the model's ability to preserve structurally consistent representations, resulting in degraded link prediction performance. This highlights the role of GSR in reinforcing the hierarchical organization learned during quantization.
(3) Quantized entity tokens provide the largest performance gains. Removing entity tokens produces the most substantial degradation demonstrating that discrete semantic units are crucial for enabling LLMs to exploit KG-grounded representations. 
\subsection{Efficiency Analysis}
\begin{figure}[t]
    \centering
    \subfigure[FB15k237]{
    \includegraphics[width=0.48\linewidth]{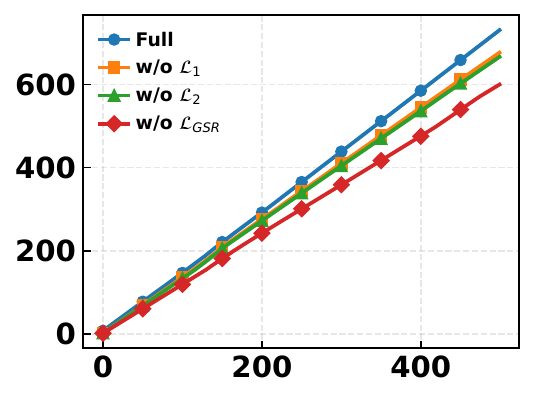}}
    \subfigure[WN18RR]{
    \includegraphics[width=0.48\linewidth]{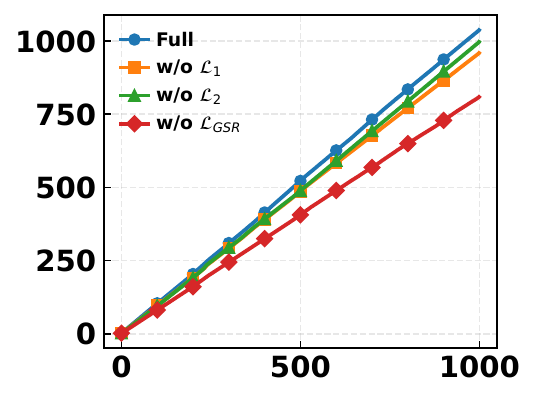}}
    \caption{
    Visualization of cumulative training time.
    }
    \label{fig:efficiency_combined}
\end{figure}

To quantify the computational overhead of each key component, we measure the total training time on both datasets and summarize the results in a single figure. As shown in Figure~\ref{fig:efficiency_combined}, removing any individual component yields only a modest reduction in cost: eliminating $\mathcal{L}_1$ or $\mathcal{L}_2$ shortens training by 4\%, while removing GSR results in a larger but still moderate 18\% decrease. These findings show that the quantization-related objectives introduce only limited computational overhead. Detailed training time statistics are reported in Appendix~\ref{sec:efficiency_details}.
\subsection{Hyperparameter Analysis}

We analyze the sensitivity of our model to four key hyperparameters: $\lambda_{1}$, $\lambda_{2}$, $\lambda_{s}$, and $\lambda_{h}$.
We vary each value within a designated range while keeping others fixed. The results show that a lower $\lambda_{s}$ and a higher $\lambda_{h}$ generally yield better performance. 
The best results appear when $\lambda_{1}=0.8$ and $\lambda_{2}=0.4$, indicating that the model benefits from a balanced configuration of these regularization terms.
Overall, our model shows robustness to $\lambda_{1}$ and $\lambda_{2}$: by maintaining these parameters within the appropriate range, it achieves competitive performance.
\begin{figure}[t]
    \centering
    \subfigure[$\lambda_{1}$]{
    \includegraphics[width=0.48\linewidth]{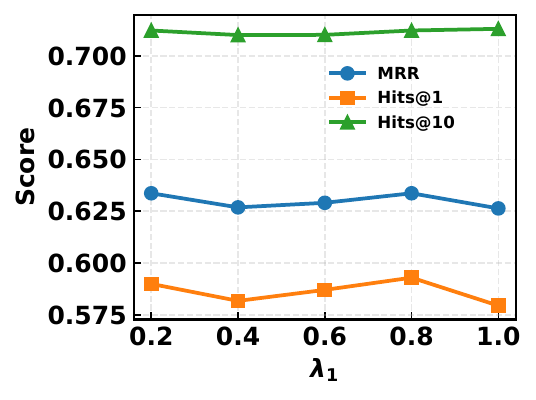}}
    \subfigure[$\lambda_{2}$]{
    \includegraphics[width=0.48\linewidth]{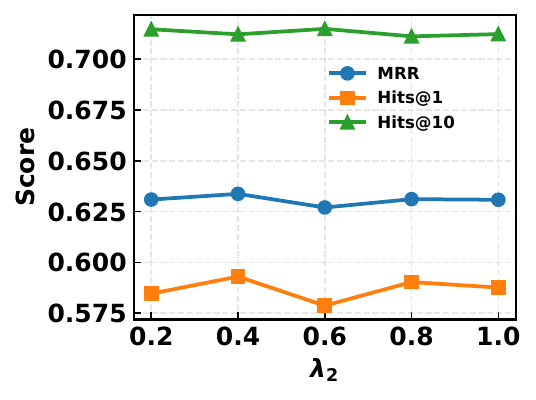}}
    \subfigure[$\lambda_{s}$]{
    \includegraphics[width=0.48\linewidth]{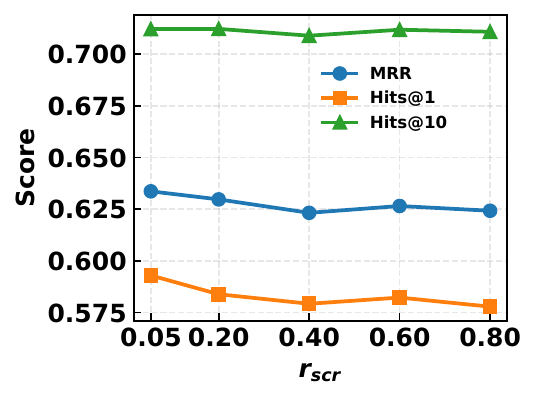}}
    \subfigure[$\lambda_{h}$]{
    \includegraphics[width=0.48\linewidth]{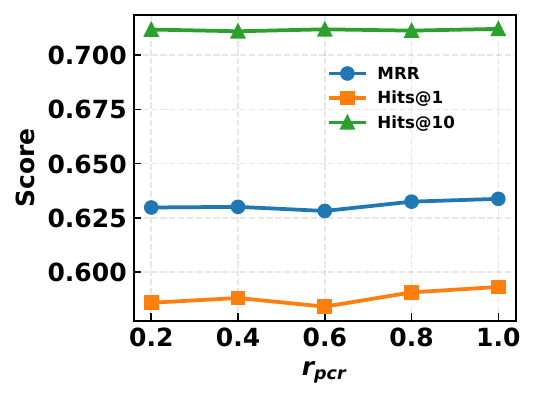}}
    \caption{Parameter sensitivity analysis on WN18RR.}
    \label{fig:params_analysis}
\end{figure}

\subsection{Codebook Analysis}
\begin{figure}[t]
    \centering
    \subfigure[FB15k237 (Ours)]{
    \includegraphics[width=0.48\linewidth]
    {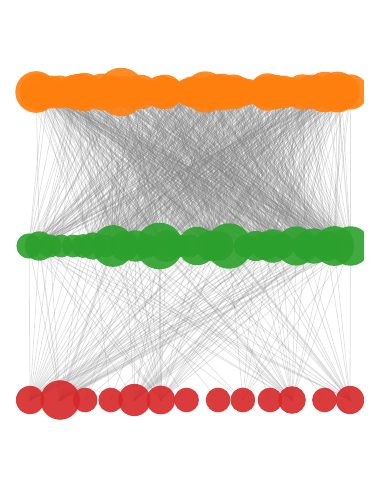}}
    \subfigure[FB15k237 (Vanilla)]{
    \includegraphics[width=0.48\linewidth]{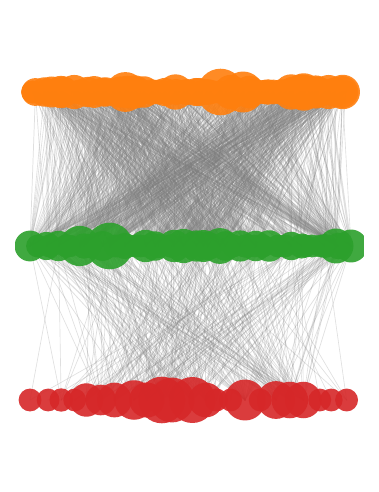}}
    \caption{Visualization of the layer distribution of entity codes.}
    \label{fig:codebook_analysis}
\end{figure}

Figure~\ref{fig:codebook_analysis} illustrates the semantic hierarchy, comparing our method against a vanilla baseline implemented as a standard RQ-VAE\cite{RQVAE}. Specifically, in this visualization, nodes colored red, green, and orange correspond to discrete codes within the coarse-, medium-, and fine-grained quantization layers. The size of each node indicates its connectivity degree, reflecting the usage frequency of the code. The edges linking nodes across layers demonstrate how distinct codes are hierarchically composed to form specific entity representations.

As shown in the figure, the coarse-grained codes exhibit a sparse yet globally uniform distribution, while the fine-grained layers display a significantly more discriminative distribution. This observation indicates that our model successfully disentangles semantic granularity, effectively correcting the misalignment observed in the baseline where broad concepts and specific entities are often conflated.

\subsection{Codebook Entropy Analysis}
\label{sec:entropy_analysis}
\paragraph{Correlation between Entropy and Performance.}
To validate our selection strategy, we analyze the relationship between $\mathcal{Y}$ and downstream performance metrics on the FB15k-237 and WN18RR dataset. In Table~\ref{tab:entropy_vs_perf}, a higher codebook entropy $\mathcal{Y}$ consistently correlates with better performance across MRR and Hits@K metrics. This positive correlation confirms that $\mathcal{Y}$ serves as a robust indicator for identifying the most effective model checkpoint.

\begin{table}[t]
\centering
\setlength{\tabcolsep}{1.5pt}
\resizebox{\linewidth}{!}{
\begin{tabular}{lcccc}
\toprule
\textbf{Entropy ($\mathcal{Y}$)} & \textbf{MRR} & \textbf{Hits@1} & \textbf{Hits@3} & \textbf{Hits@10} \\

\midrule
\multicolumn{5}{c}{\textbf{FB15k-237}} \\
\midrule
1.2285 & 0.447 \dec{0.8} & 0.376 \dec{1.0} & 0.473 \dec{0.6} & 0.585 \dec{0.7} \\
1.7339 & 0.451 \dec{0.4} & 0.381 \dec{0.5} & 0.477 \dec{0.2} & 0.589 \dec{0.3} \\
1.9421 & 0.453 \dec{0.2} & 0.384 \dec{0.3} & 0.477 \dec{0.2} & 0.590 \dec{0.2} \\
\textbf{2.2156} & \textbf{0.455} \pad & \textbf{0.386} \pad & \textbf{0.479} \pad & \textbf{0.592} \pad \\

\midrule
\multicolumn{5}{c}{\textbf{WN18RR}} \\
\midrule
0.1909 & 0.629 \dec{0.6} & 0.583 \dec{1.2} & 0.646 \dec{0.3} & 0.711 \dec{0.0} \\
0.6229 & 0.633 \dec{0.2} & 0.588 \dec{0.7} & 0.646 \dec{0.3} & 0.712 \dec{0.0} \\
1.2897 & 0.633 \dec{0.2} & 0.590 \dec{0.4} & 0.648 \dec{0.1} & 0.712 \dec{0.0} \\
\textbf{1.3852} & \textbf{0.635} \pad & \textbf{0.594} \pad & \textbf{0.649} \pad & \textbf{0.712} \pad \\

\bottomrule
\end{tabular}
}
\caption{Impact of codebook entropy on ranking performance. 
For each dataset, scores are annotated with the absolute gap to the best configuration.}
\label{tab:entropy_vs_perf}
\end{table}

\begin{figure}[t]
    \centering
    \subfigure[FB15k237]{
    \includegraphics[width=0.48\linewidth]
    {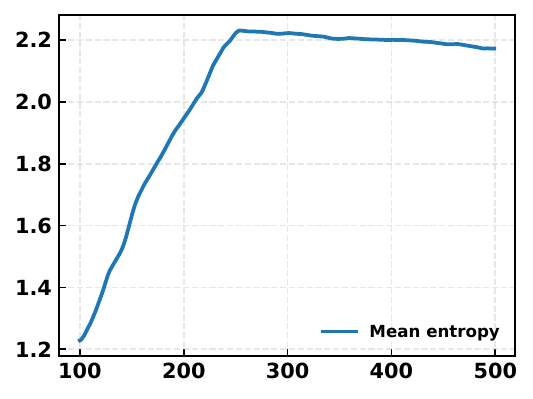}}
    \subfigure[WN18RR]{
    \includegraphics[width=0.48\linewidth]{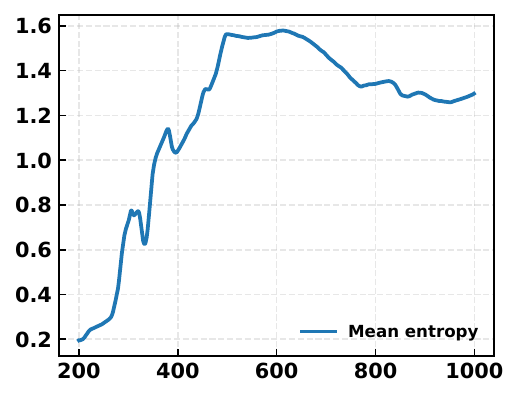}}
    \caption{Evolution of $\mathcal{Y}$ over training steps.}
    \label{fig:entropy_curves}
\end{figure}

\paragraph{Entropy Evolution During Training.}
We further visualize the evolution of codebook entropy throughout the training process. Figure~\ref{fig:entropy_curves} illustrates the changes in entropy over training steps on both FB15k-237 and WN18RR datasets. Ideally, the entropy should grow or stabilize at a high value, indicating that the model is continuously exploring and utilizing the codebook space effectively.

\section{Additional Analysis}
\label{sec:additional_analysis}
We provide comprehensive supplementary analyses of GS-Quant in the Appendix, covering: (i) training dynamics (Appendix~\ref{sec:trainning_process}); (ii) the layout of KG token embeddings (Appendix~\ref{sec:token_embedding}); (iii) qualitative case studies (Appendix~\ref{sec:case_study}); (iv) sensitivity to KG embedding initializations (Appendix~\ref{sec:init_embeddings}); (v)  robustness across varying LLM backbones (Appendix~\ref{sec:different_llms}).

\section{Conclusion}
In this paper, we presented GS-Quant, a framework bridging KGs and LLMs via quantization. By integrating Granular Semantic Enhancement for coarse-to-fine consistency and Generative Structural Reconstruction for alignment with LLM generation dynamics, GS-Quant produces semantically rich discrete codes. Extensive experiments confirm that our approach consistently outperforms state-of-the-art methods on standard benchmarks, validating the effectiveness of hierarchical semantic quantization for knowledge–LLM integration.

\clearpage

\section*{Limitations}
Despite the promising results, this work has several limitations that are worth noting.

First, our framework relies on a pre-trained large language model as the backbone. Although experiments across multiple 7B-scale models demonstrate robustness, the overall performance and efficiency are still bounded by the capacity and inference cost of the underlying LLM. Scaling the approach to larger models or deploying it in low-resource settings may introduce additional computational challenges.

Second, the proposed quantization and codebook construction are learned on specific knowledge graph benchmarks. While we observe consistent improvements on WN18RR and FB15k-237, the generalization of the learned code semantics to other knowledge graphs or domains with substantially different structural properties remains to be systematically evaluated.


Finally, our evaluation primarily focuses on link prediction tasks. While the design of GS-Quant is motivated by enhancing generative reasoning in LLMs, further validation on downstream generative or reasoning-intensive tasks would be necessary to fully assess its broader applicability.

We leave these directions for future work.

\section*{Ethical Considerations}
We adhere to the ACL Code of Ethics and have taken every measure to ensure that our research complies with the ethical guidelines set forth. Our work does not involve human subjects, nor does it raise any ethical concerns related to privacy or data usage. All datasets used in our experiments are publicly available, and we ensure that their release and use adhere to proper data-sharing policies. 
We have carefully selected the datasets and evaluation metrics to ensure fairness and transparency in our findings. No potential conflicts of interest, sponsorship biases, or ethical violations have been identified in our study.
We commit to maintaining the highest standards of research integrity, and we are open to addressing any ethical concerns that may arise during the review process.

\section*{Acknowledgments}
This work is partially supported by National Science and Technology Major Project (2026ZD1611200), and Fundamental and Interdisciplinary Disciplines Breakthrough Plan of the Ministry of Education of China (No. JYB2025XDXM118).

\bibliography{main}

\begin{thebibliography}{59}
\providecommand{\natexlab}[1]{#1}

\bibitem[{Agrawal et~al.(2024)Agrawal, Kumarage, Alghamdi, and Liu}]{KG4HalluSurvey}
Garima Agrawal, Tharindu Kumarage, Zeyad Alghamdi, and Huan Liu. 2024.
\newblock \href {https://aclanthology.org/2024.naacl-long.219/} {Can knowledge graphs reduce hallucinations in llms?: A survey}.
\newblock In \emph{Proceedings of the 2024 Conference of the North American Chapter of the Association for Computational Linguistics: Human Language Technologies (Volume 1: Long Papers)}, pages 3947--3960. Association for Computational Linguistics.

\bibitem[{Balazevic et~al.(2019)Balazevic, Allen, and Hospedales}]{TuckER}
Ivana Balazevic, Carl Allen, and Timothy Hospedales. 2019.
\newblock \href {https://doi.org/10.18653/v1/D19-1522} {{T}uck{ER}: Tensor factorization for knowledge graph completion}.
\newblock In \emph{Proceedings of the 2019 Conference on Empirical Methods in Natural Language Processing and the 9th International Joint Conference on Natural Language Processing (EMNLP-IJCNLP)}, pages 5185--5194, Hong Kong, China. Association for Computational Linguistics.

\bibitem[{Bordes et~al.(2013)Bordes, Usunier, Garcia-Dur\'{a}n, Weston, and Yakhnenko}]{TransE}
Antoine Bordes, Nicolas Usunier, Alberto Garcia-Dur\'{a}n, Jason Weston, and Oksana Yakhnenko. 2013.
\newblock Translating embeddings for modeling multi-relational data.
\newblock In \emph{Proceedings of the 27th International Conference on Neural Information Processing Systems - Volume 2}, NIPS'13, page 2787–2795.

\bibitem[{Cao et~al.(2022)Cao, Xu, Yang, Cao, and Huang}]{GIE}
Zongsheng Cao, Qianqian Xu, Zhiyong Yang, Xiaochun Cao, and Qingming Huang. 2022.
\newblock \href {https://doi.org/10.1609/aaai.v36i5.20491} {Geometry interaction knowledge graph embeddings}.
\newblock \emph{Proceedings of the AAAI Conference on Artificial Intelligence}, 36(5):5521--5529.

\bibitem[{Chami et~al.(2020)Chami, Wolf, Juan, Sala, Ravi, and R{\'e}}]{ATTH}
Ines Chami, Adva Wolf, Da-Cheng Juan, Frederic Sala, Sujith Ravi, and Christopher R{\'e}. 2020.
\newblock \href {https://doi.org/10.18653/v1/2020.acl-main.617} {Low-dimensional hyperbolic knowledge graph embeddings}.
\newblock In \emph{Proceedings of the 58th Annual Meeting of the Association for Computational Linguistics}, pages 6901--6914, Online. Association for Computational Linguistics.

\bibitem[{Chen et~al.(2020)Chen, Wang, Zhao, Cheng, Zhao, and Duan}]{KGCSurvey}
Zhe Chen, Yuehan Wang, Bin Zhao, Jing Cheng, Xin Zhao, and Zongtao Duan. 2020.
\newblock Knowledge graph completion: A review.
\newblock \emph{Ieee Access}, 8:192435--192456.

\bibitem[{Choi et~al.(2021)Choi, Jang, and Ko}]{MEM-KGC}
Bonggeun Choi, Daesik Jang, and Youngjoong Ko. 2021.
\newblock \href {https://doi.org/10.1109/ACCESS.2021.3113329} {Mem-kgc: Masked entity model for knowledge graph completion with pre-trained language model}.
\newblock \emph{IEEE Access}, 9:132025--132032.

\bibitem[{DeepSeek-AI et~al.(2025)DeepSeek-AI, Guo, Yang, Zhang, Song, Zhang, Xu, Zhu, Ma, Wang, Bi, Zhang, Yu, Wu, Wu, Gou, Shao, Li, Gao, Liu, Xue, Wang, Wu, Feng, Lu, Zhao, Deng, Zhang, Ruan, Dai, Chen, Ji, Li, Lin, Dai, Luo, Hao, Chen, Li, Zhang, Bao, Xu, Wang, Ding, Xin, Gao, Qu, Li, Guo, Li, Wang, Chen, Yuan, Qiu, Li, Cai, Ni, Liang, Chen, Dong, Hu, Gao, Guan, Huang, Yu, Wang, Zhang, Zhao, Wang, Zhang, Xu, Xia, Zhang, Zhang, Tang, Li, Wang, Li, Tian, Huang, Zhang, Wang, Chen, Du, Ge, Zhang, Pan, Wang, Chen, Jin, Chen, Lu, Zhou, Chen, Ye, Wang, Yu, Zhou, Pan, Li, Zhou, Wu, Ye, Yun, Pei, Sun, Wang, Zeng, Zhao, Liu, Liang, Gao, Yu, Zhang, Xiao, An, Liu, Wang, Chen, Nie, Cheng, Liu, Xie, Liu, Yang, Li, Su, Lin, Li, Jin, Shen, Chen, Sun, Wang, Song, Zhou, Wang, Shan, Li, Wang, Wei, Zhang, Xu, Li, Zhao, Sun, Wang, Yu, Zhang, Shi, Xiong, He, Piao, Wang, Tan, Ma, Liu, Guo, Ou, Wang, Gong, Zou, He, Xiong, Luo, You, Liu, Zhou, Zhu, Xu, Huang, Li, Zheng, Zhu, Ma, Tang, Zha, Yan, Ren, Ren, Sha, Fu, Xu, Xie, Zhang,
  Hao, Ma, Yan, Wu, Gu, Zhu, Liu, Li, Xie, Song, Pan, Huang, Xu, Zhang, and Zhang}]{deepseekr1}
DeepSeek-AI, Daya Guo, Dejian Yang, Haowei Zhang, Junxiao Song, Ruoyu Zhang, Runxin Xu, Qihao Zhu, Shirong Ma, Peiyi Wang, Xiao Bi, Xiaokang Zhang, Xingkai Yu, Yu~Wu, Z.~F. Wu, Zhibin Gou, Zhihong Shao, Zhuoshu Li, Ziyi Gao, and 181 others. 2025.
\newblock \href {https://arxiv.org/abs/2501.12948} {Deepseek-r1: Incentivizing reasoning capability in llms via reinforcement learning}.
\newblock \emph{Preprint}, arXiv:2501.12948.

\bibitem[{Guo et~al.(2024)Guo, Bo, Chen, Zhang, Chen, Yarong, Sun, Zhang, Luo, Li et~al.}]{MKGL}
Lingbing Guo, Zhongpu Bo, Zhuo Chen, Yichi Zhang, Jiaoyan Chen, Lan Yarong, Mengshu Sun, Zhiqiang Zhang, Yangyifei Luo, Qian Li, and 1 others. 2024.
\newblock Mkgl: mastery of a three-word language.
\newblock \emph{Advances in Neural Information Processing Systems}, 37:140509--140534.

\bibitem[{Guo et~al.(2020)Guo, Zhuang, Qin, Zhu, Xie, Xiong, and He}]{KG4RecSurvey}
Qingyu Guo, Fuzhen Zhuang, Chuan Qin, Hengshu Zhu, Xing Xie, Hui Xiong, and Qing He. 2020.
\newblock A survey on knowledge graph-based recommender systems.
\newblock \emph{IEEE Transactions on Knowledge and Data Engineering}, 34(8):3549--3568.

\bibitem[{Guo et~al.(2025)Guo, Wang, Chen, Guo, Li, and Chen}]{ReaLM}
Wenbin Guo, Xin Wang, Jiaoyan Chen, Lingbing Guo, Zhao Li, and Zirui Chen. 2025.
\newblock \href {https://arxiv.org/abs/2510.09711} {Realm: Residual quantization bridging knowledge graph embeddings and large language models}.
\newblock \emph{Preprint}, arXiv:2510.09711.

\bibitem[{Guo et~al.(2026)Guo, Diao, Yang, and Shi}]{GTBPE}
Zeyuan Guo, Enmao Diao, Cheng Yang, and Chuan Shi. 2026.
\newblock \href {https://openreview.net/forum?id=jCctxI1BGF} {Graph tokenization for bridging graphs and transformers}.
\newblock In \emph{The Fourteenth International Conference on Learning Representations}.

\bibitem[{Guti\'{e}rrez et~al.(2024)Guti\'{e}rrez, Shu, Gu, Yasunaga, and Su}]{HippoRAG}
Bernal~Jim\'{e}nez Guti\'{e}rrez, Yiheng Shu, Yu~Gu, Michihiro Yasunaga, and Yu~Su. 2024.
\newblock \href {https://doi.org/10.52202/079017-1902} {Hipporag: Neurobiologically inspired long-term memory for large language models}.
\newblock In \emph{Advances in Neural Information Processing Systems}, volume~37, pages 59532--59569. Curran Associates, Inc.

\bibitem[{He et~al.(2024)He, Yuan, Chen, and Horrocks}]{HiTs}
Yuan He, Moy Yuan, Jiaoyan Chen, and Ian Horrocks. 2024.
\newblock \href {https://openreview.net/forum?id=GJMYvWzjE1} {Language models as hierarchy encoders}.
\newblock In \emph{The Thirty-eighth Annual Conference on Neural Information Processing Systems}.

\bibitem[{Hu et~al.(2022)Hu, yelong shen, Wallis, Allen-Zhu, Li, Wang, Wang, and Chen}]{LoRA}
Edward~J Hu, yelong shen, Phillip Wallis, Zeyuan Allen-Zhu, Yuanzhi Li, Shean Wang, Lu~Wang, and Weizhu Chen. 2022.
\newblock \href {https://openreview.net/forum?id=nZeVKeeFYf9} {Lo{RA}: Low-rank adaptation of large language models}.
\newblock In \emph{International Conference on Learning Representations}.

\bibitem[{Jiang et~al.(2023)Jiang, Sablayrolles, Mensch, Bamford, Chaplot, de~las Casas, Bressand, Lengyel, Lample, Saulnier, Lavaud, Lachaux, Stock, Scao, Lavril, Wang, Lacroix, and Sayed}]{Mistral}
Albert~Q. Jiang, Alexandre Sablayrolles, Arthur Mensch, Chris Bamford, Devendra~Singh Chaplot, Diego de~las Casas, Florian Bressand, Gianna Lengyel, Guillaume Lample, Lucile Saulnier, Lélio~Renard Lavaud, Marie-Anne Lachaux, Pierre Stock, Teven~Le Scao, Thibaut Lavril, Thomas Wang, Timothée Lacroix, and William~El Sayed. 2023.
\newblock \href {https://arxiv.org/abs/2310.06825} {Mistral 7b}.
\newblock \emph{Preprint}, arXiv:2310.06825.

\bibitem[{Jiang et~al.(2024)Jiang, Cao, Xiao, Bhatia, Sun, and Han}]{KGFIT}
Pengcheng Jiang, Lang Cao, Cao Xiao, Parminder Bhatia, Jimeng Sun, and Jiawei Han. 2024.
\newblock \href {https://doi.org/10.52202/079017-4328} {Kg-fit: Knowledge graph fine-tuning upon open-world knowledge}.
\newblock In \emph{Advances in Neural Information Processing Systems}, volume~37, pages 136220--136258. Curran Associates, Inc.

\bibitem[{Lee et~al.(2022)Lee, Kim, Kim, Cho, and Han}]{RQVAE}
Doyup Lee, Chiheon Kim, Saehoon Kim, Minsu Cho, and Wook-Shin Han. 2022.
\newblock \href {https://doi.org/10.1109/CVPR52688.2022.01123} {Autoregressive image generation using residual quantization}.
\newblock In \emph{2022 IEEE/CVF Conference on Computer Vision and Pattern Recognition (CVPR)}, pages 11513--11522.

\bibitem[{Li et~al.(2024)Li, Tan, Chen, and Liu}]{CD}
Dawei Li, Zhen Tan, Tianlong Chen, and Huan Liu. 2024.
\newblock \href {https://aclanthology.org/2024.findings-eacl.32/} {Contextualization distillation from large language model for knowledge graph completion}.
\newblock In \emph{Findings of the Association for Computational Linguistics: EACL 2024}, pages 458--477. Association for Computational Linguistics.

\bibitem[{Li et~al.(2025)Li, Yang, Xu, Jiang, Qi, Guo, Leung, and King}]{KGR3}
Muzhi Li, Cehao Yang, Chengjin Xu, Xuhui Jiang, Yiyan Qi, Jian Guo, Ho-fung Leung, and Irwin King. 2025.
\newblock \href {https://doi.org/10.18653/v1/2025.naacl-long.221} {Retrieval, reasoning, re-ranking: A context-enriched framework for knowledge graph completion}.
\newblock In \emph{Proceedings of the 2025 Conference of the Nations of the Americas Chapter of the Association for Computational Linguistics: Human Language Technologies (Volume 1: Long Papers)}, pages 4349--4363.

\bibitem[{Lin et~al.(2025)Lin, Zhao, He, Peng, Xu, Huang, Ma, and Feng}]{SSQR}
Qika Lin, Tianzhe Zhao, Kai He, Zhen Peng, Fangzhi Xu, Ling Huang, Jingying Ma, and Mengling Feng. 2025.
\newblock \href {https://doi.org/10.18653/v1/2025.acl-long.667} {Self-supervised quantized representation for seamlessly integrating knowledge graphs with large language models}.
\newblock In \emph{Proceedings of the 63rd Annual Meeting of the Association for Computational Linguistics (Volume 1: Long Papers)}, pages 13587--13602, Vienna, Austria. Association for Computational Linguistics.

\bibitem[{Liu et~al.(2025{\natexlab{a}})Liu, Zhang, Lin, Yang, and Peng}]{FtG}
Ben Liu, Jihai Zhang, Fangquan Lin, Cheng Yang, and Min Peng. 2025{\natexlab{a}}.
\newblock \href {https://aclanthology.org/2025.coling-main.740/} {Filter-then-generate: Large language models with structure-text adapter for knowledge graph completion}.
\newblock In \emph{Proceedings of the 31st International Conference on Computational Linguistics}, pages 11181--11195.

\bibitem[{Liu et~al.(2022)Liu, Sun, Li, and Hu}]{CoLE}
Yang Liu, Zequn Sun, Guangyao Li, and Wei Hu. 2022.
\newblock \href {https://doi.org/10.1145/3511808.3557355} {I know what you do not know: Knowledge graph embedding via co-distillation learning}.
\newblock In \emph{Proceedings of the 31st ACM International Conference on Information \& Knowledge Management}, CIKM '22, page 1329–1338, New York, NY, USA. Association for Computing Machinery.

\bibitem[{Liu et~al.(2024)Liu, Tian, Sun, and Hu}]{DIFT}
Yang Liu, Xiaobin Tian, Zequn Sun, and Wei Hu. 2024.
\newblock \href {https://doi.org/10.1007/978-3-031-77844-5_11} {Finetuning generative large language models with discrimination instructions for knowledge graph completion}.
\newblock In \emph{The Semantic Web – ISWC 2024: 23rd International Semantic Web Conference, Baltimore, MD, USA, November 11–15, 2024, Proceedings, Part I}, page 199–217, Berlin, Heidelberg. Springer-Verlag.

\bibitem[{Liu et~al.(2025{\natexlab{b}})Liu, Cao, Lin, Shang, Wang, and Pan}]{SAT}
Yu~Liu, Yanan Cao, Xixun Lin, Yanmin Shang, Shi Wang, and Shirui Pan. 2025{\natexlab{b}}.
\newblock \href {https://doi.org/10.18653/v1/2025.emnlp-main.1061} {Enhancing large language model for knowledge graph completion via structure-aware alignment-tuning}.
\newblock In \emph{Proceedings of the 2025 Conference on Empirical Methods in Natural Language Processing}, pages 20981--20995.

\bibitem[{Ma et~al.(2025)Ma, Chen, Wu, Khan, and Wang}]{LLM4KGQASurvey}
Chuangtao Ma, Yongrui Chen, Tianxing Wu, Arijit Khan, and Haofen Wang. 2025.
\newblock \href {https://doi.org/10.18653/v1/2025.emnlp-main.1249} {Large language models meet knowledge graphs for question answering: Synthesis and opportunities}.
\newblock In \emph{Proceedings of the 2025 Conference on Empirical Methods in Natural Language Processing}, pages 24589--24608, Suzhou, China. Association for Computational Linguistics.

\bibitem[{Ma et~al.(2023)Ma, Song, Hu, Li, Zhang, and King}]{ConcreteGraph}
Yueen Ma, Zixing Song, Xuming Hu, Jingjing Li, Yifei Zhang, and Irwin King. 2023.
\newblock Graph component contrastive learning for concept relatedness estimation.
\newblock In \emph{Proceedings of the AAAI Conference on Artificial Intelligence}, pages 13362--13370.

\bibitem[{M{\"u}llner(2011)}]{agglomerativeclustering}
Daniel M{\"u}llner. 2011.
\newblock Modern hierarchical, agglomerative clustering algorithms.
\newblock \emph{arXiv preprint arXiv:1109.2378}.

\bibitem[{Nathani et~al.(2019)Nathani, Chauhan, Sharma, and Kaul}]{KBGAT}
Deepak Nathani, Jatin Chauhan, Charu Sharma, and Manohar Kaul. 2019.
\newblock \href {https://doi.org/10.18653/v1/P19-1466} {Learning attention-based embeddings for relation prediction in knowledge graphs}.
\newblock In \emph{Proceedings of the 57th Annual Meeting of the Association for Computational Linguistics}, pages 4710--4723, Florence, Italy. Association for Computational Linguistics.

\bibitem[{Ouyang et~al.(2022)Ouyang, Wu, Jiang, Almeida, Wainwright, Mishkin, Zhang, Agarwal, Slama, Ray, Schulman, Hilton, Kelton, Miller, Simens, Askell, Welinder, Christiano, Leike, and Lowe}]{SFT}
Long Ouyang, Jeff Wu, Xu~Jiang, Diogo Almeida, Carroll~L. Wainwright, Pamela Mishkin, Chong Zhang, Sandhini Agarwal, Katarina Slama, Alex Ray, John Schulman, Jacob Hilton, Fraser Kelton, Luke Miller, Maddie Simens, Amanda Askell, Peter Welinder, Paul Christiano, Jan Leike, and Ryan Lowe. 2022.
\newblock Training language models to follow instructions with human feedback.
\newblock In \emph{Proceedings of the 36th International Conference on Neural Information Processing Systems}, NIPS '22, Red Hook, NY, USA. Curran Associates Inc.

\bibitem[{Pan et~al.(2024)Pan, Luo, Wang, Chen, Wang, and Wu}]{KGLLM}
Shirui Pan, Linhao Luo, Yufei Wang, Chen Chen, Jiapu Wang, and Xindong Wu. 2024.
\newblock \href {https://doi.org/10.1109/TKDE.2024.3352100} {Unifying large language models and knowledge graphs: A roadmap}.
\newblock \emph{IEEE Transactions on Knowledge and Data Engineering}, pages 1--20.

\bibitem[{Peng et~al.(2024)Peng, Zhu, Liu, Bo, Shi, Hong, Zhang, and Tang}]{GraphRAG}
Boci Peng, Yun Zhu, Yongchao Liu, Xiaohe Bo, Haizhou Shi, Chuntao Hong, Yan Zhang, and Siliang Tang. 2024.
\newblock Graph retrieval-augmented generation: A survey.
\newblock \emph{ACM Transactions on Information Systems}.

\bibitem[{Schlichtkrull et~al.(2018)Schlichtkrull, Kipf, Bloem, Van Den~Berg, Titov, and Welling}]{RGCN}
Michael Schlichtkrull, Thomas~N Kipf, Peter Bloem, Rianne Van Den~Berg, Ivan Titov, and Max Welling. 2018.
\newblock Modeling relational data with graph convolutional networks.
\newblock In \emph{European semantic web conference}, pages 593--607. Springer.

\bibitem[{Song et~al.(2022)Song, Ma, and King}]{NewFrontiersinGraphLearning}
Zixing Song, Yueen Ma, and Irwin King. 2022.
\newblock Individual fairness in dynamic financial networks.
\newblock In \emph{NeurIPS 2022 Workshop: New Frontiers in Graph Learning}.

\bibitem[{Song et~al.(2024)Song, Meng, and King}]{DBLP:conf/aaai/SongMK24}
Zixing Song, Ziqiao Meng, and Irwin King. 2024.
\newblock A diffusion-based pre-training framework for crystal property prediction.
\newblock In \emph{{AAAI}}, pages 8993--9001. {AAAI} Press.

\bibitem[{Song et~al.(2023{\natexlab{a}})Song, Zhang, and King}]{DBLP:conf/nips/SongZK23}
Zixing Song, Yifei Zhang, and Irwin King. 2023{\natexlab{a}}.
\newblock No change, no gain: Empowering graph neural networks with expected model change maximization for active learning.
\newblock In \emph{NeurIPS}.

\bibitem[{Song et~al.(2023{\natexlab{b}})Song, Zhang, and King}]{DBLP:conf/nips/SongZK23a}
Zixing Song, Yifei Zhang, and Irwin King. 2023{\natexlab{b}}.
\newblock Optimal block-wise asymmetric graph construction for graph-based semi-supervised learning.
\newblock In \emph{NeurIPS}.

\bibitem[{Sun et~al.(2024)Sun, Xu, Tang, Wang, Lin, Gong, Ni, Shum, and Guo}]{ToG}
Jiashuo Sun, Chengjin Xu, Lumingyuan Tang, Saizhuo Wang, Chen Lin, Yeyun Gong, Lionel Ni, Heung-Yeung Shum, and Jian Guo. 2024.
\newblock \href {https://openreview.net/forum?id=nnVO1PvbTv} {Think-on-graph: Deep and responsible reasoning of large language model on knowledge graph}.
\newblock In \emph{The Twelfth International Conference on Learning Representations}.

\bibitem[{Sun et~al.(2019)Sun, Deng, Nie, and Tang}]{RotatE}
Zhiqing Sun, Zhi-Hong Deng, Jian-Yun Nie, and Jian Tang. 2019.
\newblock Rotate: Knowledge graph embedding by relational rotation in complex space.
\newblock In \emph{International Conference on Learning Representations}.

\bibitem[{Tenenbaum et~al.(2011)Tenenbaum, Kemp, Griffiths, and Goodman}]{tenenbaum2011grow}
Joshua~B Tenenbaum, Charles Kemp, Thomas~L Griffiths, and Noah~D Goodman. 2011.
\newblock \href {https://www.science.org/doi/10.1126/science.1192788} {How to grow a mind: Statistics, structure, and abstraction}.
\newblock \emph{science}, 331(6022):1279--1285.

\bibitem[{Trouillon et~al.(2016)Trouillon, Welbl, Riedel, Gaussier, and Bouchard}]{Complex}
Théo Trouillon, Johannes Welbl, Sebastian Riedel, Eric Gaussier, and Guillaume Bouchard. 2016.
\newblock \href {https://proceedings.mlr.press/v48/trouillon16.html} {Complex embeddings for simple link prediction}.
\newblock In \emph{Proceedings of The 33rd International Conference on Machine Learning}, volume~48 of \emph{Proceedings of Machine Learning Research}, pages 2071--2080, New York, New York, USA. PMLR.

\bibitem[{van~den Oord et~al.(2017)van~den Oord, Vinyals, and kavukcuoglu}]{VQVAE}
Aaron van~den Oord, Oriol Vinyals, and koray kavukcuoglu. 2017.
\newblock \href {https://proceedings.neurips.cc/paper_files/paper/2017/file/7a98af17e63a0ac09ce2e96d03992fbc-Paper.pdf} {Neural discrete representation learning}.
\newblock In \emph{Advances in Neural Information Processing Systems}, volume~30. Curran Associates, Inc.

\bibitem[{van~der Maaten and Hinton(2008)}]{tsne}
Laurens van~der Maaten and Geoffrey Hinton. 2008.
\newblock \href {http://jmlr.org/papers/v9/vandermaaten08a.html} {Visualizing data using t-sne}.
\newblock \emph{Journal of Machine Learning Research}, 9(86):2579--2605.

\bibitem[{Vashishth et~al.(2020)Vashishth, Sanyal, Nitin, and Talukdar}]{CompGCN}
Shikhar Vashishth, Soumya Sanyal, Vikram Nitin, and Partha Talukdar. 2020.
\newblock \href {https://openreview.net/forum?id=BylA_C4tPr} {Composition-based multi-relational graph convolutional networks}.
\newblock In \emph{International Conference on Learning Representations}.

\bibitem[{Wang et~al.(2025)Wang, Fan, Feng, Shanru, Ma, Wang, and Yin}]{K-RagRec}
Shijie Wang, Wenqi Fan, Yue Feng, Lin Shanru, Xinyu Ma, Shuaiqiang Wang, and Dawei Yin. 2025.
\newblock \href {https://doi.org/10.18653/v1/2025.acl-long.1317} {Knowledge graph retrieval-augmented generation for {LLM}-based recommendation}.
\newblock In \emph{Proceedings of the 63rd Annual Meeting of the Association for Computational Linguistics (Volume 1: Long Papers)}, pages 27152--27168, Vienna, Austria. Association for Computational Linguistics.

\bibitem[{Wang et~al.(2021)Wang, Gao, Zhu, Zhang, Liu, Li, and Tang}]{KEPLER}
Xiaozhi Wang, Tianyu Gao, Zhaocheng Zhu, Zhengyan Zhang, Zhiyuan Liu, Juanzi Li, and Jian Tang. 2021.
\newblock \href {https://doi.org/10.1162/tacl_a_00360} {{KEPLER: A Unified Model for Knowledge Embedding and Pre-trained Language Representation}}.
\newblock \emph{Transactions of the Association for Computational Linguistics}, 9:176--194.

\bibitem[{Wei et~al.(2023)Wei, Huang, Zhang, and Kwok}]{KICGPT}
Yanbin Wei, Qiushi Huang, Yu~Zhang, and James Kwok. 2023.
\newblock \href {https://doi.org/10.18653/v1/2023.findings-emnlp.580} {{KICGPT}: Large language model with knowledge in context for knowledge graph completion}.
\newblock In \emph{Findings of the Association for Computational Linguistics: EMNLP 2023}, pages 8667--8683, Singapore. Association for Computational Linguistics.

\bibitem[{Xu et~al.(2024)Xu, Zhang, Lin, Wu, Zhu, Xu, Zhao, Zheng, and Chen}]{MPIKGC}
Derong Xu, Ziheng Zhang, Zhenxi Lin, Xian Wu, Zhihong Zhu, Tong Xu, Xiangyu Zhao, Yefeng Zheng, and Enhong Chen. 2024.
\newblock \href {https://aclanthology.org/2024.lrec-main.1044/} {Multi-perspective improvement of knowledge graph completion with large language models}.
\newblock In \emph{Proceedings of the 2024 Joint International Conference on Computational Linguistics, Language Resources and Evaluation (LREC-COLING 2024)}, pages 11956--11968.

\bibitem[{Yang et~al.(2024)Yang, Yang, Hui, Zheng, Yu, Zhou, Li, Li, Liu, Huang, Dong, Wei, Lin, Tang, Wang, Yang, Tu, Zhang, Ma, Yang, Xu, Zhou, Bai, He, Lin, Dang, Lu, Chen, Yang, Li, Xue, Ni, Zhang, Wang, Peng, Men, Gao, Lin, Wang, Bai, Tan, Zhu, Li, Liu, Ge, Deng, Zhou, Ren, Zhang, Wei, Ren, Liu, Fan, Yao, Zhang, Wan, Chu, Liu, Cui, Zhang, Guo, and Fan}]{Qwen2}
An~Yang, Baosong Yang, Binyuan Hui, Bo~Zheng, Bowen Yu, Chang Zhou, Chengpeng Li, Chengyuan Li, Dayiheng Liu, Fei Huang, Guanting Dong, Haoran Wei, Huan Lin, Jialong Tang, Jialin Wang, Jian Yang, Jianhong Tu, Jianwei Zhang, Jianxin Ma, and 43 others. 2024.
\newblock \href {https://arxiv.org/abs/2407.10671} {Qwen2 technical report}.
\newblock \emph{Preprint}, arXiv:2407.10671.

\bibitem[{Yang et~al.(2015)Yang, Yih, He, Gao, and Deng}]{DistMult}
Bishan Yang, Wen{-}tau Yih, Xiaodong He, Jianfeng Gao, and Li~Deng. 2015.
\newblock Embedding entities and relations for learning and inference in knowledge bases.
\newblock In \emph{3rd International Conference on Learning Representations, {ICLR} 2015, San Diego, CA, USA, May 7-9, 2015, Conference Track Proceedings}.

\bibitem[{Yang et~al.(2025)Yang, Yang, Zhu, Li, Zhang, Li, and Li}]{SLiNT}
Mengxue Yang, Chun Yang, Jiaqi Zhu, Jiafan Li, Jingqi Zhang, Yuyang Li, and Ying Li. 2025.
\newblock \href {https://doi.org/10.18653/v1/2025.findings-emnlp.736} {{SL}i{NT}: Structure-aware language model with injection and contrastive training for knowledge graph completion}.
\newblock In \emph{Findings of the Association for Computational Linguistics: EMNLP 2025}, pages 13658--13671, Suzhou, China.

\bibitem[{Yao et~al.(2019)Yao, Mao, and Luo}]{kgbert}
Liang Yao, Chengsheng Mao, and Yuan Luo. 2019.
\newblock \href {https://arxiv.org/abs/1909.03193} {Kg-bert: Bert for knowledge graph completion}.

\bibitem[{Zhang et~al.(2024)Zhang, Chen, Guo, yajing Xu, Zhang, and Chen}]{KoPA}
Yichi Zhang, Zhuo Chen, Lingbing Guo, yajing Xu, Wen Zhang, and Huajun Chen. 2024.
\newblock \href {https://openreview.net/forum?id=HHzHRuIyaW} {Making large language models perform better in knowledge graph completion}.
\newblock In \emph{ACM Multimedia 2024}.

\bibitem[{Zhang and Yao(2022)}]{RED-GNN}
Yongqi Zhang and Quanming Yao. 2022.
\newblock \href {https://doi.org/10.1145/3485447.3512008} {Knowledge graph reasoning with relational digraph}.
\newblock In \emph{Proceedings of the ACM Web Conference 2022}, WWW '22, page 912–924, New York, NY, USA. Association for Computing Machinery.

\bibitem[{Zhao et~al.(2021)Zhao, Qi, Liu, and Zhang}]{WGCN}
Yunxiang Zhao, Jianzhong Qi, Qingwei Liu, and Rui Zhang. 2021.
\newblock \href {https://doi.org/10.1145/3404835.3462834} {Wgcn: Graph convolutional networks with weighted structural features}.
\newblock In \emph{Proceedings of the 44th International ACM SIGIR Conference on Research and Development in Information Retrieval}, SIGIR '21, page 624–633, New York, NY, USA. Association for Computing Machinery.

\bibitem[{Zheng et~al.(2023)Zheng, Chiang, Sheng, Zhuang, Wu, Zhuang, Lin, Li, Li, Xing, Zhang, Gonzalez, and Stoica}]{Vicuna}
Lianmin Zheng, Wei-Lin Chiang, Ying Sheng, Siyuan Zhuang, Zhanghao Wu, Yonghao Zhuang, Zi~Lin, Zhuohan Li, Dacheng Li, Eric~P. Xing, Hao Zhang, Joseph~E. Gonzalez, and Ion Stoica. 2023.
\newblock \href {https://arxiv.org/abs/2306.05685} {Judging llm-as-a-judge with mt-bench and chatbot arena}.
\newblock \emph{Preprint}, arXiv:2306.05685.

\bibitem[{Zhou et~al.(2020)Zhou, Cui, Hu, Zhang, Yang, Liu, Wang, Li, and Sun}]{GNN}
Jie Zhou, Ganqu Cui, Shengding Hu, Zhengyan Zhang, Cheng Yang, Zhiyuan Liu, Lifeng Wang, Changcheng Li, and Maosong Sun. 2020.
\newblock \href {https://doi.org/10.1016/j.aiopen.2021.01.001} {Graph neural networks: A review of methods and applications}.
\newblock \emph{AI Open}, 1:57--81.

\bibitem[{Zhu et~al.(2023)Zhu, Yuan, Galkin, Xhonneux, Zhang, Gazeau, and Tang}]{ANet}
Zhaocheng Zhu, Xinyu Yuan, Michael Galkin, Louis-Pascal Xhonneux, Ming Zhang, Maxime Gazeau, and Jian Tang. 2023.
\newblock \href {https://proceedings.neurips.cc/paper_files/paper/2023/file/b9e98316cb72fee82cc1160da5810abc-Paper-Conference.pdf} {Aast net: A scalable path-based reasoning approach for knowledge graphs}.
\newblock In \emph{Advances in Neural Information Processing Systems}, volume~36, pages 59323--59336. Curran Associates, Inc.

\bibitem[{Zhu et~al.(2021)Zhu, Zhang, Xhonneux, and Tang}]{NBF-Net}
Zhaocheng Zhu, Zuobai Zhang, Louis-Pascal Xhonneux, and Jian Tang. 2021.
\newblock \href {https://proceedings.neurips.cc/paper_files/paper/2021/file/f6a673f09493afcd8b129a0bcf1cd5bc-Paper.pdf} {Neural bellman-ford networks: A general graph neural network framework for link prediction}.
\newblock In \emph{Advances in Neural Information Processing Systems}, volume~34.

\end{thebibliography}

\appendix

\section{Implementation Details} 
\label{sec:implementation}

The statistics of utilized datasets are shown in Table~\ref{tab:dataset}. We conduct all our experiments on a Linux server equipped with two Intel Xeon Gold 6248R processors and eight A100-PCIE-40GB GPUs.\footnote{Only a subset of the GPUs are utilized in our experiments.} 

To ensure a fair comparison among LLM-based methods, we standardize the experimental configurations for DIFT, SSQR, and our approach. Specifically, we employ Llama3-8B as the unified backbone to eliminate architectural variance and adopt the CoLE candidate sets and prompts released by DIFT. Under this aligned protocol, we reproduce the SSQR results while keeping its other hyperparameters consistent with the official implementation.

For parameter-efficient fine-tuning, we adopt LoRA~\cite{LoRA}, with the detailed prompts provided in Appendix~\ref{sec:prompts}. For hierarchical clustering, we use the LLM-enhanced agglomerative clustering following KG-FIT~\cite{KGFIT}. All related hyperparameters are reported in Table~\ref{tab:hyperparameters}. The discrete codebooks of GS-Quant are learned with a dedicated decoder-only module, whose training configurations are summarized in Table~\ref{tab:codebook_hyperparameters}. The computational cost of LoRA-based re-ranking is presented in Table~\ref{tab:timecomplexity}.

\begin{table}[htbp]\centering
\small
\begin{tabular}{ccc}
\toprule
\textbf{Dataset}& \textbf{FB15k237} & \textbf{WN18RR}\\
\midrule
Entities & 14,541 & 40,943 \\
Relations & 237 & 11 \\
Train & 272,115 & 86,835 \\
Valid & 17,535 & 3,034 \\
Test & 20,466 & 3,134 \\
\bottomrule
\end{tabular}
\caption{Statistics of Datasets}
\label{tab:dataset}
\vspace{-0.15cm}
\end{table}

\begin{table}[t]\centering
\small
\begin{tabular}{lc}
\toprule
\textbf{Hyperparameters} & \textbf{Settings} \\
\midrule
Precision & bf16 \\
Random seed & 42 \\
Learning rate & 2e-4 \\

Per-device train batch size & 16 \\
Per-device eval batch size & 8 \\
Gradient accumulation steps & 1 \\
Optimizer & AdamW \\
Scheduler & Constant LR \\
Warmup ratio & 0.03 \\
Weight decay & 0.0 \\
Max grad norm & 1.0 \\
DeepSpeed config & Zero-3 \\
LoRA rank & 32 \\
LoRA $\alpha$ & 64 \\
Max new tokens & 64 \\
Min new tokens & 1 \\
Max training steps & 3800 \\
\midrule
\multicolumn{2}{c}{\textbf{Dataset-Specific Hyperparameters}} \\
\midrule
\textbf{FB15K-237} &
\begin{tabular}[t]{@{}l@{}}
Max training epoch: 8 \\
\end{tabular} \\
\midrule
\textbf{WN18RR} &
\begin{tabular}[t]{@{}l@{}}
Max training epoch: 4 \\
\end{tabular} \\
\bottomrule
\end{tabular}
\caption{Hyperparameters for LoRA and LLM}
\label{tab:hyperparameters}
\end{table}

\begin{table}[htbp]\centering
\small
\begin{tabular}{lc}
\toprule
\textbf{Hyperparameters} & \textbf{Settings} \\
\midrule
Embedding dim & 512 \\
Codebook layers & 4 \\
Codebook size & 1024 \\
Encoder layers & [512, 512, 512] \\
Reconstruction layers & 2 \\
Reconstruction heads & 4 \\
Parent recon count & 5 \\
Dropout probability & 0.0 \\
$\lambda_1$ & 0.8 \\
$\lambda_2$ & 0.4 \\
Learning rate & 1e{-4} \\
Commit loss weight & 0.25 \\
\midrule
\multicolumn{2}{c}{\textbf{Dataset-Specific Hyperparameters}} \\
\midrule
\textbf{FB15K-237} &
\begin{tabular}[t]{@{}l@{}}
Max steps: 500 \\
Batch size: 14{,}541 \\
$\lambda_{s}$: 1 \\
$\lambda_{h}$: 1
\end{tabular} \\
\midrule
\textbf{WN18RR} &
\begin{tabular}[t]{@{}l@{}}
Max steps: 1000 \\
Batch size: 16{,}348 \\
$\lambda_{s}$: 0.05 \\
$\lambda_{h}$: 1
\end{tabular} \\
\bottomrule
\end{tabular}
\caption{Hyperparameters for training the quantized codebooks.}
\label{tab:codebook_hyperparameters}
\end{table}

\section{Training Process of GS-Quant}
\label{sec:trainning_process}
\begin{figure}[t]
    \centering
    \subfigure[FB15k237]{
    \includegraphics[width=0.48\linewidth]
    {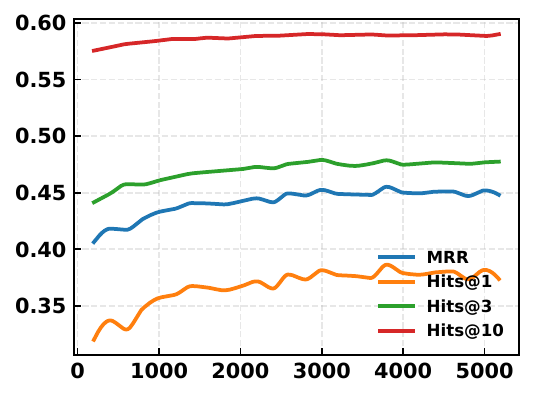}}
    \subfigure[WN18RR]{
    \includegraphics[width=0.48\linewidth]{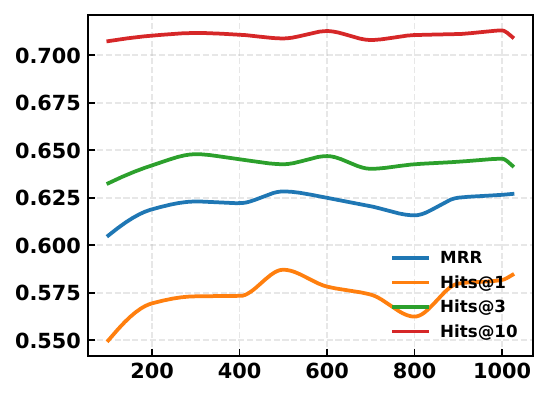}}
    \caption{Visualization of the evolution of MRR and Hits@\{1,3,10\} across training checkpoints.}
    \label{fig:codebook_training_analysis}
\end{figure}

Figure~\ref{fig:codebook_training_analysis} presents the training process of GS-Quant on FB15K-237 and WN18RR, reporting the evolution of MRR and Hits@\{1,3,10\} across training checkpoints. Overall, GS-Quant shows stable convergence behavior on both datasets, with consistent improvements across all evaluation metrics.

On WN18RR, the model converges rapidly in the early stage, where MRR and Hits@1 increase noticeably within the first few hundred checkpoints, and then stabilize with only minor fluctuations. The best performance is reached around checkpoint 1024, indicating that GS-Quant can efficiently capture the core semantic structure of this dataset.

In contrast, FB15K-237 exhibits a longer optimization process, with performance improving steadily over a larger number of training steps. The model achieves its peak MRR at checkpoint 3800, while maintaining stable performance in later stages. This difference in training steps is mainly due to the substantially larger training set of FB15K-237 compared to WN18RR.

Notably, among all metrics, Hits@1 shows the most significant relative improvement during training on both datasets. The gains in Hits@1 and MRR best highlight the practical value of our approach, as they directly reflect the model's ability to rank the correct entity at the top position and provide the correct answer immediately, rather than merely placing it within a broader candidate set.

\section{Detailed Training Time Statistics}
\label{sec:efficiency_details}
This section reports detailed wall-clock training time statistics for GS-Quant and its ablated variants. 
All experiments are conducted under identical hardware configurations and training settings, with only the corresponding objective terms removed for each variant.

Table~\ref{tab:train_cost} summarizes the total training time on FB15k-237 and WN18RR, as well as the absolute and relative time savings compared to the full model. 
These statistics provide a detailed reference for the efficiency trends discussed in the main text.
\begin{table}[t]
\centering
\small
\setlength{\tabcolsep}{4pt}
\begin{tabular}{l|cc|cc}
\toprule
& \multicolumn{2}{c|}{\textbf{FB15K-237}} 
& \multicolumn{2}{c}{\textbf{WN18RR}} \\
\textbf{Settings} 
& \textbf{Time (s)} & \textbf{Saved (\%)} 
& \textbf{Time (s)} & \textbf{Saved (\%)} \\
\midrule
\textbf{Full} 
& 730.48 & --- 
& 1037.34 & --- \\
w/o $\mathcal{L}_1$ 
& 676.27 & 7.42\% 
& 959.00 & 7.55\% \\
w/o $\mathcal{L}_2$ 
& 666.18 & 8.80\% 
& 997.03 & 3.89\% \\
w/o $\mathcal{L}_{\text{GSR}}$ 
& 599.62 & 17.91\% 
& 808.86 & 22.03\% \\
\bottomrule
\end{tabular}
\caption{Training cost comparison across ablated variants.}
\label{tab:train_cost}
\end{table}

\section{Token Embeddings in LLMs}
\label{sec:token_embedding}
\begin{figure}[t]
    \centering
    \subfigure[Ours]{
    \includegraphics[width=0.48\linewidth]{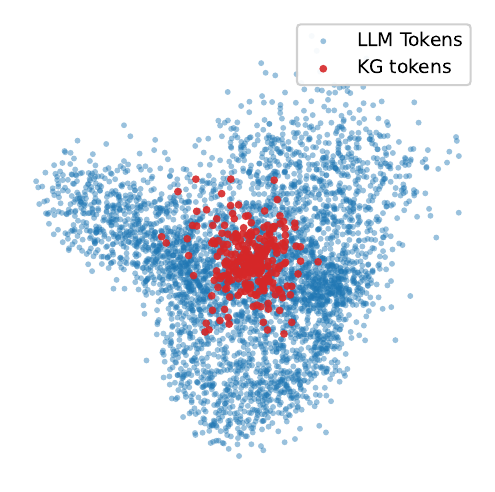}}
    \subfigure[DIFT]{
    \includegraphics[width=0.48\linewidth]{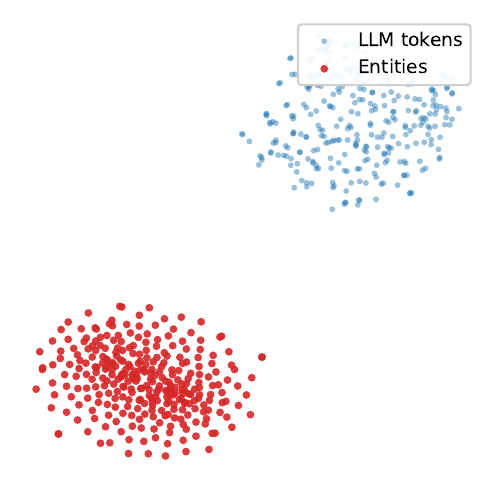}}
    \caption{Embedding virtualization in LLMs (FB15k-237 dataset): KG tokens versus base word tokens for ours, and entity embeddings versus base tokens for DIFT.}
    \label{fig:codebook_embeddding}
\end{figure}

To analyze how GS-Quant represents KG tokens in comparison with base language tokens, 
we visualize all KG tokens together with LLM tokens on the FB15k-237 dataset.
All token embeddings are projected to two dimensions using t-SNE~\cite{tsne}, as shown in Figure~\ref{fig:codebook_embeddding}.

As illustrated in Figure~\ref{fig:codebook_embeddding}(a), 
KG tokens in GS-Quant form a compact and distinguishable region that remains well integrated with the overall LLM token space.
In contrast, DIFT exhibits a more isolated separation between entity embeddings and language tokens (Figure~\ref{fig:codebook_embeddding}(b)), 
suggesting weaker embedding-space integration.
Overall, these results indicate that GS-Quant achieves token-level differentiation while preserving compatibility with the native LLM embedding geometry.

\section{The Layer Distribution on WN18RR}
\label{sec:layer_distribution}
\begin{figure}[t]
    \centering
    \subfigure[WN18RR (Ours)]{
    \includegraphics[width=0.48\linewidth]
    {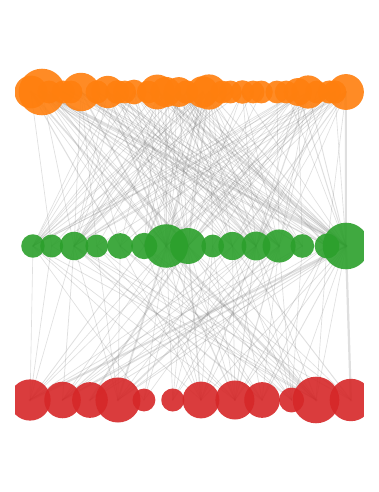}}
    \subfigure[WN18RR (Vanilla)]{
    \includegraphics[width=0.48\linewidth]
    {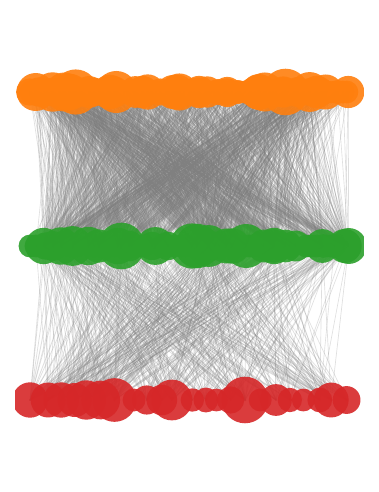}}
    \caption{Visualization of the layer distribution of entity codes.}
    \label{fig:codebook_analysis_WN18RR}
\end{figure}

As shown in Figure~\ref{fig:codebook_analysis_WN18RR}, the coarse-grained code layers maintain a sparse and globally uniform distribution, while deeper layers gradually evolve into highly discriminative and structured patterns. The evidence also suggests that our model enforces a clearer separation of semantic granularity on WN18RR, where entities with subtle relational differences are more cleanly differentiated in fine-grained layers.

\section{Case Studies}
\label{sec:case_study}

To demonstrate the robustness of our codebook-based representation, we present a qualitative analysis of four distinct cases in Table~\ref{tab:four_cases}. These examples highlight how the learned discrete codes capture multi-granular semantics to correct various types of baseline errors. In this comparison, the baseline refers to the model's configuration where the discrete codes are removed from the input prompt, forcing the model to rely solely on textual descriptions.

\paragraph{Geographical Reasoning (Case 1 \& 2).} 
The first two cases illustrate improvements in geographical reasoning. In Case 1, the baseline suffers from a "self-loop" error, failing to understand that \textit{Beverly Hills} is contained within \textit{Los Angeles}. Our model resolves this by capturing the hierarchical relationship encoded in the suffix codes. Similarly, in Case 2, the baseline predicts a "Congressional District" when the relation requires a "State" level entity. Our approach successfully corrects this granularity mismatch, identifying \textit{Virginia} as the correct administrative level.

\paragraph{Fine-grained \& Type Consistency (Case 3 \& 4).} 
The benefits extend beyond geography. Case 3 shows that our model can distinguish between fine-grained academic degrees (Bachelor vs. Master), whereas the baseline retrieves a generic related concept. Finally, Case 4 demonstrates strict type enforcement: while the baseline erroneously predicts an event (\textit{2010 Winter Olympics}) for a query about medals, our model correctly retrieves the object \textit{Bronze medal}, ensuring semantic consistency with the relation.

\begin{table}[t]
    \centering
    \small
    \renewcommand{\arraystretch}{1.3}
    \setlength{\tabcolsep}{4pt}
    \begin{tabular}{p{0.95\linewidth}}
        \toprule
        \multicolumn{1}{c}{\textbf{Case 1: Geographical Containment (Hierarchy)}} \\
        \textbf{Query:} $(?, \texttt{/location/.../contains}, \text{Beverly Hills})$ \\
        \textbf{Gold:} \textbf{Los Angeles} \\
        \ding{55} \textbf{Base:} Beverly Hills \textit{(Self-loop Error)} \\
        \ding{51} \textbf{Ours:} \textbf{Los Angeles} \\
        \emph{\textbf{Analysis:} Baseline fails to model the containment hierarchy (Part vs. Whole); Ours identifies the super-region.} \\
        \midrule
        
        \multicolumn{1}{c}{\textbf{Case 2: Administrative Granularity (Level)}} \\
        \textbf{Query:} $(?, \texttt{/location/.../religion}, \text{Presbyterianism})$ \\
        \textbf{Gold:} \textbf{Virginia} (State) \\
        \ding{55} \textbf{Base:} Pennsylvania's 12th congressional district \textit{(Too Specific)} \\
        \ding{51} \textbf{Ours:} \textbf{Virginia} \\
        \emph{\textbf{Analysis:} Baseline confuses a \underline{District} with a \underline{State}; Ours fixes the granularity to the correct admin level.} \\
        \midrule
        
        \multicolumn{1}{c}{\textbf{Case 3: Educational Degree (Fine-grained)}} \\
        \textbf{Query:} $(?, \texttt{/.../institution}, \text{Royal Holloway})$ \\
        \textbf{Gold:} \textbf{Bachelor of Arts} \\
        \ding{55} \textbf{Base:} Master of Arts \textit{(Wrong Degree Level)} \\
        \ding{51} \textbf{Ours:} \textbf{Bachelor of Arts} \\
        \emph{\textbf{Analysis:} Baseline retrieves a related but incorrect degree; Ours distinguishes fine-grained academic ranks.} \\
        \midrule
        
        \multicolumn{1}{c}{\textbf{Case 4: Entity Type Consistency (Semantic)}} \\
        \textbf{Query:} $(\text{Bohemia}, \texttt{/olympics/.../medal}, ?)$ \\
        \textbf{Gold:} \textbf{Bronze medal} \\
        \ding{55} \textbf{Base:} 2010 Winter Olympics \textit{(Event vs. Object)} \\
        \ding{51} \textbf{Ours:} \textbf{Bronze medal} \\
        \emph{\textbf{Analysis:} Baseline predicts an event instead of a medal object; Ours enforces strict type consistency.} \\
        \bottomrule
    \end{tabular}
    \caption{Qualitative examples comparing our codebook-based approach against the baseline across four distinct scenarios. The codebook effectively corrects errors related to hierarchical containment, administrative level alignment, fine-grained semantic discrimination, and entity type consistency.}
    \label{tab:four_cases}
\end{table}
\vspace{-5pt}

\section{Comparison of Different KG Embeddings}
\label{sec:init_embeddings}
To demonstrate the robustness of our framework, we employ four representative models as initial embeddings: ComplEx~\cite{Complex}, DistMult~\cite{DistMult}, pRotatE~\cite{RotatE}, and TransE~\cite{TransE}. 
The quantitative comparison is reported in Table~\ref{tab:wn18rr_init_embeddings}. It is worth noting that the results for the baselines (Base) and the competitor KG-FIT~\cite{KGFIT} are collected from their original literature and our reproduction experiments to ensure a fair comparison.

As shown in the table, our framework achieves consistent and substantial improvements over the original embedding models across all metrics. Furthermore, when compared to KG-FIT, our method maintains a clear advantage, particularly in terms of MRR and Hits@1. While KG-FIT shows competitive performance in broader retrieval metrics, our framework demonstrates superior ranking precision, effectively placing the correct entities at the top ranks. These results highlight the strong generalization capacity of our framework, regardless of the specific KG embedding model used.
\begin{table}[t]
\centering
\footnotesize  
\setlength{\tabcolsep}{3pt} 
\renewcommand{\arraystretch}{0.95} 
\begin{tabular}{l|l|cccc}
\toprule
\textbf{Model} & \textbf{Setting} & \textbf{MRR} & \textbf{Hits@1} & \textbf{Hits@3} & \textbf{Hits@10} \\
\midrule
\multirow{5}{*}{ComplEx} 
 & Base   & 0.440 & 0.410 & 0.460 & 0.510 \\
 & KG-FIT & 0.585 & 0.520 & - & 0.705 \\
 & \textbf{Ours} & \textbf{0.630} & \textbf{0.588} & \textbf{0.646} & \textbf{0.710} \\
 & \textit{\scriptsize $\Delta$ vs Base} & \textit{\scriptsize +0.190} & \textit{\scriptsize +0.178} & \textit{\scriptsize +0.186} & \textit{\scriptsize +0.200} \\
 & \textit{\scriptsize $\Delta$ vs FIT}  & \textit{\scriptsize +0.045} & \textit{\scriptsize +0.068} & - & \textit{\scriptsize +0.005} \\
\midrule
\multirow{5}{*}{DistMult} 
 & Base   & 0.430 & 0.390 & 0.440 & 0.490 \\
 & KG-FIT & 0.590 & 0.530 & - & 0.710 \\
 & \textbf{Ours} & \textbf{0.636} & \textbf{0.595} & \textbf{0.651} & \textbf{0.713} \\
 & \textit{\scriptsize $\Delta$ vs Base} & \textit{\scriptsize +0.206} & \textit{\scriptsize +0.205} & \textit{\scriptsize +0.211} & \textit{\scriptsize +0.223} \\
 & \textit{\scriptsize $\Delta$ vs FIT}  & \textit{\scriptsize +0.046} & \textit{\scriptsize +0.065} & - & \textit{\scriptsize +0.003} \\
\midrule
\multirow{5}{*}{pRotatE} 
 & Base   & 0.460 & 0.416 & 0.479 & 0.552 \\
 & KG-FIT & 0.590 & 0.511 & - & 0.722 \\
 & \textbf{Ours} & \textbf{0.626} & \textbf{0.581} & \textbf{0.646} & \textbf{0.709} \\
 & \textit{\scriptsize $\Delta$ vs Base} & \textit{\scriptsize +0.166} & \textit{\scriptsize +0.165} & \textit{\scriptsize +0.167} & \textit{\scriptsize +0.157} \\
 & \textit{\scriptsize $\Delta$ vs FIT}  & \textit{\scriptsize +0.036} & \textit{\scriptsize +0.070} & - & \textit{\scriptsize -0.013} \\
\midrule
\multirow{5}{*}{TransE} 
 & Base   & 0.243 & 0.043 & 0.441 & 0.532 \\
 & KG-FIT & 0.550 & 0.450 & - & 0.695 \\
 & \textbf{Ours} & \textbf{0.621} & \textbf{0.570} & \textbf{0.647} & \textbf{0.713} \\
 & \textit{\scriptsize $\Delta$ vs Base} & \textit{\scriptsize +0.378} & \textit{\scriptsize +0.527} & \textit{\scriptsize +0.206} & \textit{\scriptsize +0.181} \\
 & \textit{\scriptsize $\Delta$ vs FIT}  & \textit{\scriptsize +0.071} & \textit{\scriptsize +0.120} & - & \textit{\scriptsize +0.018} \\
\bottomrule
\end{tabular}
\caption{Detailed performance comparison on WN18RR. ``$\Delta$'' denotes the absolute performance gain of our method compared to the baselines and KG-FIT. ``-'' indicates the metric was not reported in the baseline paper.}
\label{tab:wn18rr_init_embeddings}
\end{table}
\vspace{-5pt}

\section{Performance with different LLMs}\label{sec:different_llms}
This appendix experiment evaluates the robustness of our framework under different LLM backbones, including Mistral~\cite{Mistral}, Qwen2.5~\cite{Qwen2}, Vicuna~\cite{Vicuna}, and DeepSeek-R1-Distill-Qwen~\cite{deepseekr1}. 
All models are trained and evaluated using the same data splits, optimization settings, and hyperparameters. 
As shown in Table~\ref{tab:llm_results}, the performance remains stable across different LLM choices on both WN18RR and FB15k-237, indicating that GS-Quant does not rely on a specific LLM implementation and exhibits strong robustness to backbone variations.

\begin{table}[htbp]
\begin{tabular}{lcc}
\toprule
\textbf{Dataset} & \textbf{SFT Time} & \textbf{Re-rank Time} \\
\midrule
FB15K-237 & 1h 56min & 0.180s / query \\
WN18RR    & 17min    & 0.152s / query \\
\bottomrule
\end{tabular}
\caption{Training and inference time costs for SFT and re-ranking.}
\label{tab:timecomplexity}
\end{table}
\vspace{-5pt}

\section{Prompt Templates}\label{sec:prompts}
Table~\ref{tab:pt_gq} presents the prompt templates employed in our method for knowledge graph completion.
\begin{table*}[t]
\setlength{\tabcolsep}{3pt}
\centering
\begin{tabular}{lcccccccc}
\toprule
\multirow{2}{*}{\parbox[c]{3.2cm}{\centering \textbf{Model}}}
& \multicolumn{4}{c}{\textbf{WN18RR}}
& \multicolumn{4}{c}{\textbf{FB15k-237}} \\
\cmidrule(lr){2-5} \cmidrule(lr){6-9}
& \textbf{MRR} & \textbf{Hits@1} & \textbf{Hits@3} & \textbf{Hits@10}
& \textbf{MRR} & \textbf{Hits@1} & \textbf{Hits@3} & \textbf{Hits@10} \\
\midrule

Qwen2.5-14B
& \textbf{0.6292} & \textbf{0.5842} & 0.6496 & 0.7117
& 0.4506 & 0.3799 & 0.4761 & 0.5888 \\

Qwen2.5-7B
& \underline{0.6249} & \underline{0.5747} & \textbf{0.6517} & \textbf{0.7128}
& 0.4385 & \underline{0.3640} & 0.4654 & 0.5860 \\

Mistral-7B
& 0.6205 & 0.5723 & \underline{0.6431} & \underline{0.7123}
& \textbf{0.4515} & \textbf{0.3810} & \textbf{0.4774} & \underline{0.5891} \\

Llama-2-13B
& 0.6082 & 0.5585 & 0.6335 & 0.7068
& 0.4465 & 0.3722 & 0.4750 & 0.5885 \\

Llama-2-7B
& 0.6105 & 0.5628 & 0.6295 & 0.7058
& 0.4378 & 0.3615 & 0.4672 & 0.5855 \\

Vicuna-13B
& 0.6065 & 0.5547 & 0.6305 & 0.7056
& \underline{0.4484} & 0.3750 & \underline{0.4771} & \textbf{0.5900} \\

Vicuna-7B
& 0.6121 & 0.5645 & 0.6310 & 0.7055
& 0.4391 & 0.3635 & 0.4687 & 0.5866 \\

DeepSeek-R1-Distill-Qwen-7B
& 0.6010 & 0.5455 & 0.6286 & 0.7042
& 0.4189 & 0.3362 & 0.4512 & 0.5797 \\

\bottomrule
\end{tabular}
\caption{Comparison of LLM-based KGC performance on WN18RR and FB15k-237.
The best results are \textbf{boldfaced} and the second-best are \underline{underlined}.}
\label{tab:llm_results}
\end{table*}

\begin{table*}[!ht]
\centering
\begin{tabularx}{\textwidth}{X}
\rule{\textwidth}{1.5pt}
\textcolor{darkblue}{\textbf{\#\# Query triple}}
(telephone, \textit{verb group}, < ??? >)
\\
\rule{\textwidth}{1.5pt}
\vspace{0.6em}
\textcolor{darkblue}{\textbf{\#\# Prompts}}

\textbf{User}: Here is a triplet with tail entity $t$ unknown:
(telephone, \textit{verb group}, $t$).

\vspace{0.6em}

Following are some details about $t$:

telephone, get or try to get into communication (with someone) by telephone;
``I tried to call you all night'';
``Take two aspirin and call me in the morning''

\vspace{0.6em}

Quantized representation:

\texttt{<\#bau><\#ya><\#bcq><\#rm>}

\vspace{0.6em}

Following are some triplets about $t$:

(telephone, derivationally related form, tintinnabulation);
(telephone, derivationally related form, telephony);
(telephone, synset domain topic of, telephony);
(telephone, derivationally related form, telephoner);
(telephone, hypernym, telecommunicate);
(telephone, derivationally related form, telephone call);
(tintinnabulation, derivationally related form, telephone);
(telephoner, derivationally related form, telephone);
(telephony, derivationally related form, telephone);
(telephone call, derivationally related form, telephone)

\vspace{0.6em}

What is the entity name of $t$? Select one from the list:

\vspace{0.4em}

\begin{tabular}{@{}l l@{}}
call \texttt{<\#bau><\#ri><\#ci><\#ux>} &
telephone \texttt{<\#bau><\#ya><\#bcq><\#rm>} \\

cell phone \texttt{<\#bau><\#ri><\#yz><\#bbp>} &
call in \texttt{<\#bau><\#ya><\#bcq><\#cg>} \\

telephonist \texttt{<\#bau><\#ri><\#bfy><\#nx>} &
telephony \texttt{<\#bau><\#bn><\#yz><\#s>} \\

telephone call \texttt{<\#bau><\#bdp><\#nu><\#tn>} &
telephone set \texttt{<\#bau><\#bn><\#ci><\#to>} \\

dial \texttt{<\#bau><\#ya><\#gx><\#bma>} &
telecommunicate \texttt{<\#bau><\#ya><\#tp><\#cs>} \\

telephoner \texttt{<\#bau><\#vq><\#va><\#dd>} &
netmail \texttt{<\#bau><\#ri><\#ft><\#rt>} \\

hold the line \texttt{<\#bau><\#ya><\#bw><\#bcy>} &
telex \texttt{<\#bau><\#ya><\#gx><\#wp>} \\

telefax \texttt{<\#bau><\#ya><\#gx><\#cs>} &
wire \texttt{<\#bau><\#ya><\#gx><\#mc>} \\

radiotelephonic \texttt{<\#bau><\#ya><\#bcq><\#ff>} &
trunk call \texttt{<\#bau><\#bn><\#nu><\#kk>} \\

sampling frequency \texttt{<\#bau><\#jp><\#gx><\#vw>} &
sampling \texttt{<\#bau><\#jp><\#bcq><\#vw>}
\end{tabular}
\\ \rule{\textwidth}{1.5pt}
\textcolor{darkblue}{\textbf{\#\# Results}}
\textbf{Assistant}: call \\
\rule{\textwidth}{1.5pt}
\end{tabularx}
\caption{Prompt Template for Knowledge Graph Completion.}
\label{tab:pt_gq}
\end{table*}

\end{document}